\documentclass[letterpaper]{article} 
\usepackage{aaai25}  
\nocopyright
\usepackage{times}  
\usepackage{helvet}  
\usepackage{courier}  
\usepackage[hyphens]{url}  
\usepackage{graphicx} 
\usepackage{natbib}  
\usepackage{caption} 
\usepackage{algorithm}
\usepackage{algorithmic}
\usepackage{newfloat}
\usepackage{listings}
\DeclareCaptionStyle{ruled}{labelfont=normalfont,labelsep=colon,strut=off} 
\floatstyle{ruled}
\newfloat{listing}{tb}{lst}{}
\floatname{listing}{Listing}
\usepackage{appendix}
\usepackage{amsmath,amsfonts}
\usepackage{array}
\usepackage[caption=false,font=footnotesize,labelfont=rm,textfont=rm]{subfig}
\usepackage{textcomp}
\usepackage{verbatim}
\usepackage{cite}
\usepackage{multirow}
\usepackage{booktabs}
\usepackage{makecell} 
\usepackage{subfig}
\usepackage{bbding}

\graphicspath{{imgs/}}
\title{Lifting Scheme-Based Implicit Disentanglement of\\ Emotion-Related Facial Dynamics in the Wild}
\author{
    Xingjian Wang\textsuperscript{\rm 1}, 
    Li Chai\textsuperscript{\rm 1}\thanks{Corresponding author.}
}
\affiliations{
        \textsuperscript{\rm 1}The State Key Laboratory of Industrial Control Technology, Zhejiang University\\
        \{xingjianwang, chaili\}@zju.edu.cn
}

\begin{document}

\maketitle

\begin{abstract}
In-the-wild dynamic facial expression recognition (DFER) encounters a significant challenge in recognizing emotion-related expressions, which are often temporally and spatially diluted by emotion-irrelevant expressions and global context.
Most prior DFER methods directly utilize coupled spatiotemporal representations that may incorporate weakly relevant features with emotion-irrelevant context bias.
Several DFER methods highlight dynamic information for DFER, but following explicit guidance that may be vulnerable to irrelevant motion.
In this paper, we propose a novel Implicit Facial Dynamics Disentanglement framework (IFDD).
Through expanding wavelet lifting scheme to fully learnable framework, IFDD disentangles emotion-related dynamic information from emotion-irrelevant global context in an implicit manner, i.e., without exploit operations and external guidance.
The disentanglement process contains two stages.
The first is Inter-frame Static-dynamic Splitting Module (ISSM) for rough disentanglement estimation, which explores inter-frame correlation to generate content-aware splitting indexes on-the-fly.
We utilize these indexes to split frame features into two groups, one with greater global similarity, and the other with more unique dynamic features.
The second stage is Lifting-based Aggregation-Disentanglement Module (LADM) for further refinement.
LADM first aggregates two groups of features from ISSM to obtain fine-grained global context features by an updater, and then disentangles emotion-related facial dynamic features from the global context by a predictor.
Extensive experiments on in-the-wild datasets have demonstrated that IFDD outperforms prior supervised DFER methods with higher recognition accuracy and comparable efficiency.
Code is available at \url{https://github.com/CyberPegasus/IFDD}.
\end{abstract}

\section{Introduction}
Dynamic facial expression recognition (DFER) for in-the-wild scenarios holds significant importance for understanding human mental state, and facilitate various relevant applications\cite{nature01}.
Despite the impressive progress both in static facial expression recognition (SFER) and laboratory-controlled DFER, in-the-wild DFER remains challenging.
As revealed in prior works\cite{NR-DFERNet, FreqHD}, videos of in-the-wild DFER contain limited dynamic expression-related frames along with numerous neutral or noisy frames.
That is, the emotion-related frames are temporally diluted by non-expression frames.

Previous DFER methods tend to model tightly coupled spatiotemporal representations.
They utilize the whole representation for recognition, which may be vulnerable to excessive non-expression frames in non-neutral emotion classification tasks.
Specifically, a large part of previous methods\cite{DFEW2020,STT,MAEDFER} directly apply the entire hidden features or global tokens from 3D convolutional neural network (CNN) or vision transformer (ViT) for DFER.
Besides, there are some methods\cite{Former-DETR,DPCNet,IAL} that utilize separable spatiotemporal architecture, i.e., 2D CNN + GRU/LSTM/Transformer, to extract spatial and temporal representation in a cascaded manner.
Their spatial representations are tangled with temporal ones in the end.
Thus, the above two kinds of methods accord with the mentioned issue.
Their expression representation may contain numerous emotion-irrelevant features such as facial shapes and identity-specific characteristics, leading to emotion-irrelevant context bias and information redundancy.
In light of this, we endeavor to disentangle emotion-related dynamic features from global context features for more compact and effective representation of dynamic expression.

Recently, several DFER studies\cite{NR-DFERNet,FreqHD} have responded to this challenge and focused on the significant dynamic information.
NR-DFERNet\cite{NR-DFERNet} directly differential operation between adjacent frames, thereby paying more attention to the key dynamic frames.
Freq-HD\cite{FreqHD} utilizes Discrete Fourier Transform-based (DFT) frequency analysis sliding across frames to strengthen high-dynamics affective clips.

However, their introduced explicit guidance to extract dynamics will be vulnerable to background motion and head movements, especially for in-the-wild complex scenarios.
Instead, we distill emotion-related dynamics from higher-level latent features via expanding wavelet lifting scheme \cite{WaveletLifting} to a more implicit and adaptive framework.

Given the aforementioned concerns, we propose Implicit Facial Dynamics Disentanglement (IFDD) compatible with different backbones. IFDD implicitly disentangles emotion-related dynamic features from global context features via a two-stage framework, i.e., Inter-frame Static-dynamic Splitting Module (ISSM) for rough estimation and Lifting-based Aggregation-Disentanglement Module (LADM) for further refinement.
The implicitness is reflected in two parts, i.e., without the explicit operation such as inter-frame difference\cite{NR-DFERNet} or frequency analysis\cite{FreqHD}, and without external guidance such as optical flows\cite{explicit-opticalFlow} or facial landmarks\cite{explicit-facialLandmarks}.

Different from prior works, IFDD expands wavelet lifting scheme in two parts: (1) Instead of fixed even-odd splitting, ISSM is designed as an adaptive splitting method based on temporal correlation. 
(2) Instead of solely considering single-group features for updater/predictor, LADM introduces mutual relation of two groups for supplement.

Specifically, ISSM dynamically generates content-aware indices adapting to temporal correlation of frame features.
Leveraging indices for interpolation, ISSM preliminarily splits features into two groups along the temporal dimension.
The relatively static group contains larger spatial similarity to all the frames, while the dynamic group contains unique dynamic features within a local temporal region.

Subsequently, LADM aggregates these two groups to obtain low-frequency context features by an updater, and utilizes global context loss to force the global context be incorporated.
After that, LADM calculates the correlation between context features and the dynamic group, and subtracts the global information from dynamic group by a predictor for disentanglement, resulting in emotion-related dynamic features.
Task-specific loss is applied to dynamic features.

Our main contributions are summarized as follows:
\begin{itemize}
\item We rethink previous frameworks that introduce explicit guidance to focus on dynamic information for DFER, and propose a novel implicit disentanglement framework named IFDD.
IFDD disentangles emotion-related dynamic features from the global context to alleviate the negative impact of emotion-irrelevant frames.
\item We propose the disentanglement process as a wavelet lifting-based two-stage refinement, and expand lifting scheme into a fully learnable framework.
Firstly, ISSM preliminarily splits spatiotemporal features into static and dynamic groups based on their temporal correlation.
Secondly, LADM first integrates two groups to obtain low-frequency global context information, and then utilizes global context to purify high-frequency emotion-related dynamic features from dynamic groups.
\item We integrate IFDD framework with CNN and ViT backbones.
Extensive experimental results on three in-the-wild DFER datasets demonstrate the superiority of IFDD over other state-of-the-art supervised methods.
\end{itemize}

\section{Related Work}
\subsection{DFER in the Wild}
Despite significant progress have been made in SFER\cite{SFER2,SFER3},
DFER still remains challenging since it needs to consider inter-frame temporal relationship in addition to the spatial information.
Jiang \emph{et al.} \cite{DFEW2020} and Wang \emph{et al.} \cite{FERV39k} combine 2D convolution neural network (CNN) for spatial features with recurrent network for temporal dynamics, while NR-DFERNet\cite{NR-DFERNet}, DPCNet\cite{DPCNet}, and IAL\cite{IAL} utilize transformer as the substitute for RNN.
Meanwhile, Jiang \emph{et al.} explores several 3D CNN baselines, including C3D\cite{C3D}, 3D ResNet-18\cite{3DResNet}, R(2+1)D-18\cite{R21D}, I3D\cite{I3D}, and P3D\cite{P3D}.
M3DFEL\cite{RethinkM3DFEL} equips 3D CNN with BiLSTM to further model the imbalanced temporal relationships.
To exploit transformer for 3D representation, STT\cite{STT} uses CNN to embed frames, and then jointly learns 3D representation by spatial and temporal attentions within transformer blocks.

However, their coupled spatiotemporal representations will inevitably incorporate noisy and emotion-irrelevant context information from prevalent non-expression frames of in-the-wild DFER tasks. 
Therefore, we manage to disentangle emotion-relevant and compact dynamic features from relatively irrelevant global context features.

\subsection{Tackling Noisy Non-Expression Frames of DFER}
Recently, several methods have managed to alleviate the negative impact of numerous non-expression frames in DFER from multiple perspectives, mainly including dynamic representation\cite{NR-DFERNet,FreqHD} and learning strategy\cite{RethinkM3DFEL,SeekingUncertainty}.
To enhance emotion-relevant dynamic representation, NR-DFERNet\cite{NR-DFERNet} utilizes inter-frame differential operation to introduce frame-level dynamic information, while Freq-HD\cite{FreqHD} utilizes DFT-based frequency analysis sliding across frames to focus on the key dynamic frames.
From the perspective of learning strategy, M3DFEL\cite{RethinkM3DFEL} treats this issue as a weakly supervised problem and utilize multi-instance learning to handle inexact labels, while SCIU\cite{SeekingUncertainty} manages to prune low-quality data samples in training process.

In this paper, we focus on modeling dynamic representation along with other concerns.
Prior arts\cite{NR-DFERNet,FreqHD} explicitly introduce prior knowledge to enhance dynamic representation at the frame level.
Thus, they have limited adaptive capability to complex DFER scenarios.
Instead, we rethink this idea and manage to propose an implicit, interpretable and more adaptive framework.

\subsection{Deep Learning-based Wavelet Lifting Methods}
Extending the scope of wavelet, lifting scheme\cite{WaveletLifting} has shown its promising performance in vision applications.
Given the input 1D signal $x$, the lifting scheme extracts its frequency sub-bands as detailed and approximate coefficients by three steps:
(1) Split signal in two groups including odd samples $x_o$ and even samples $x_e$.
(2) Utilize a predictor $\mathcal{P}$ to compute an estimation for odd samples $\hat{x}_o$ based on even samples, and subtract the prediction from odd samples to obtain high-frequency sub-band $h$.
(3) Utilize an updater $\mathcal{U}$ to recalibrate even samples with predicted high-frequency components to obtain low-frequency sub-band $g$.

Since lifting scheme allows for flexible customized filters, there emerges various studies on deep learning-based lifting methods.
DAWN \cite{lifting-deep} uses CNN for updater and predictor to classify images, while LGLFormer \cite{lifting-lglformer} applies self-attention to lifting framework for remote sensing scene parsing.
Via GCN-based predictor and updater, Huang \emph{et al.} \cite{lifting-3D} decomposes object shapes into sub-bands for 3D shape representation learning.
Besides, PIRNet \cite{PIRNet} utilizes the restoration property of lifting scheme for privacy-preserving image restoration.

Despite significant progress has been achieved for 2D image tasks, it is still challenging to expand lifting scheme to DFER video tasks, because of additional temporal information and noisy non-expression frames. 
Therefore, in this paper, we expand prior lifting scheme both for splitting process and predicting/updating process.

\begin{figure*}[t]
    \centering
    \includegraphics[width=0.95\textwidth]{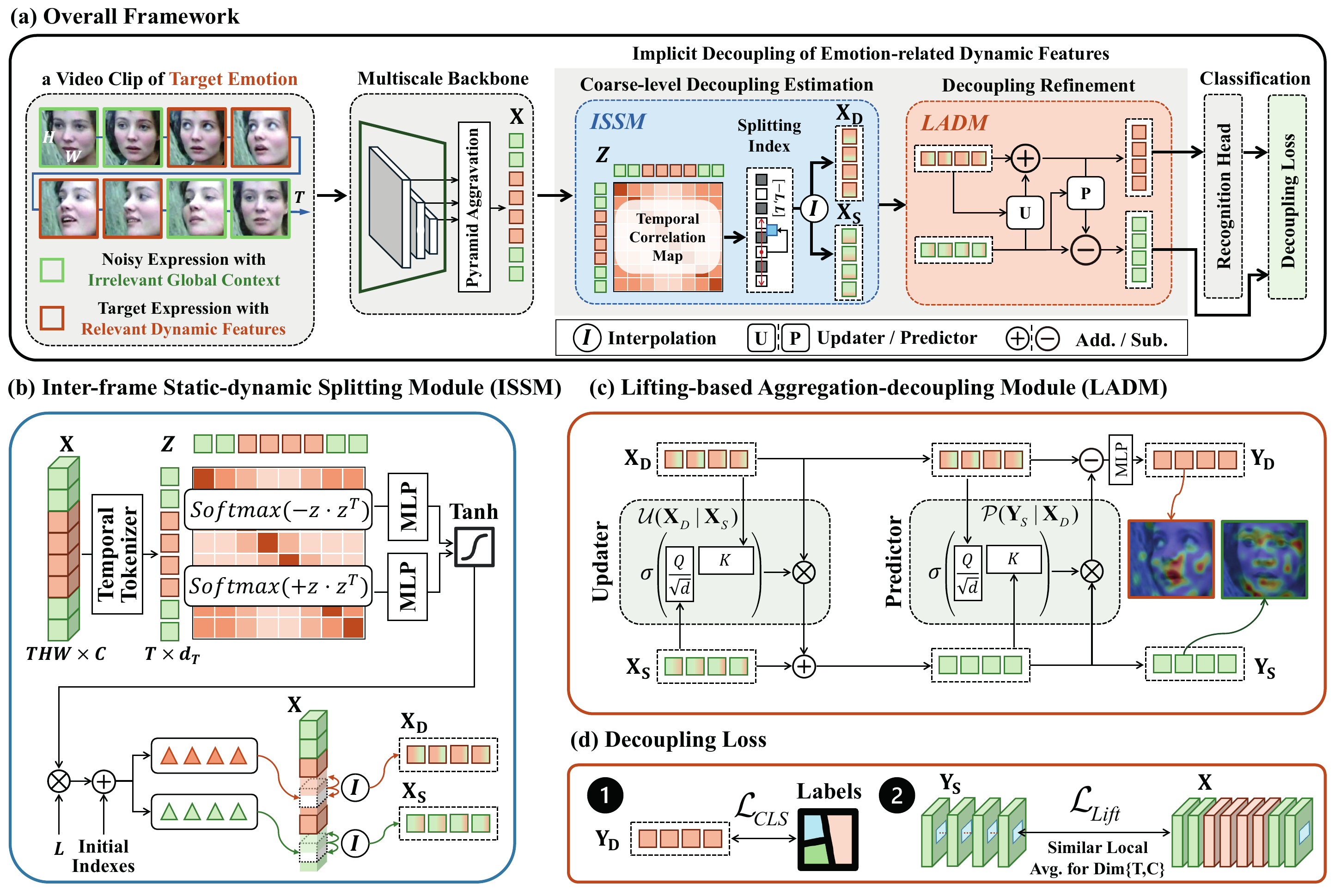}
    \caption{
            Overview framework of IFDD, which mainly consists of four parts: (1) multiscale backbone followed by pyramid aggregation; (2) Inter-frame Static-dynamic Splitting Module (ISSM); (3) Lifting-based Aggregation-Decoupling Module (LADM); (4) recognition head with decoupling loss. 
    Based on the spatiotemporal features extracted by backbone, ISSM and LADM modules are proposed to further decouple emotion-related dynamic features from emotion-irrelevant global context.
    }
    \label{Fig. Framework}
\end{figure*}

\section{Implicit Facial Dynamics Disentanglement}

The overall architecture of IFDD is shown in Fig. \ref{Fig. Framework}.
Given an input video clip $\mathbf{F}\in\mathbb{R}^{T_0\times H_0\times W_0\times 3}$ of $T_0$ RGB frames with height $H_0$ and width $W_0$, IFDD first extract multiscale features from $\mathbf{F}$ by the backbone.
Then pyramid aggregation utilizes dilated convolutions to compress multiscale features into the same spatial size, and concatenate them followed by convolution layers for aggregation, resulting in single-scale latent features $\mathbf{X}\in\mathbb{R}^{T\times H\times W\times C}$.
Subsequently, the proposed ISSM and LADM jointly conduct two-stage disentanglement of emotion-related dynamics on $\mathbf{X}$, which will be illustrated in detail in the following subsections.

\subsection{Inter-Frame Static-Dynamic Splitting}
Fixed even-odd splitting strategy is commonly used in prior methods \cite{lifting-deep,lifting-3D,PIRNet}, regarding even and odd coefficients as the preliminary estimation for low-frequency and high-frequency information respectively\cite{WaveletLifting}.
However, this is excessively rough for DFER tasks, since emotion-related expressions are non-uniformly diluted and disrupted by irrelevant expressions.

We propose Inter-frame Static-dynamic Splitting Module (ISSM) to provide a better preliminary estimation and facilitate subsequent refinement process in LADM.
ISSM generates content-aware splitting index adaptive to $\mathbf{X}$, and use these indices to split $\mathbf{X}$ into global context group and dynamic group.
The splitting indices are learned by temporal correlation to avoid being exposed to the whole spatiotemporal features with excessive emotion-irrelevant global content.
In this way, frame features in global context group exhibit higher spatial similarity to other frames across the time and thus are relatively static, while the ones of dynamic group possess unique dynamics that may indicate emotion-related expressions and thus deserve special attention.

Given spatiotemporal latent features $\mathbf{X} \in\mathbb{R}^{T\times H\times W\times C}$, ISSM first embeds $\mathbf{X}$ into temporal tokens $\mathbf{Z}\in\mathbb{R}^{T\times d_T}$ with temporal dimension $d_T$ as follows,
\begin{equation}
    \label{Eq. TemporalToken}
    \resizebox{0.9\hsize}{!}{$
    \mathbf{Z} = \mathrm{LN}\left(\mathop{Vec}\limits_{dim=2,3,4}(\mathrm{Conv}(\mathrm{P}_{S\downarrow 2} (\mathbf{X})))\cdot \mathbf{W}_T+\mathbf{b}_T\right),$}
\end{equation}
where $\cdot$ denotes matrix multiplication in this paper, $\mathcal{P}_{S\downarrow 2}$ is average pooling operation for spatial compression with a downsampling factor $2$, and $\mathop{Vec}$ denotes vectorization operator to flatten a tensor for specific dimensions $dim$.
$\mathrm{Conv}(\cdot)$ denotes convolution layer with kernel = 3 to compress the channel dimension by a factor of $\frac{1}{D_c}$, while $\mathrm{LN}(\cdot)$ represents Layer Normalization operation.
$\mathbf{W}_T\in \mathbb{R}^{\frac{HWC}{4 D_c}\times d_T}$ and $\mathbf{b}_T \in \mathbb{R}^{T\times d_T}$ denotes learnable embedding parameters.

Then ISSM learns two sets of splitting indices based on the temporal correlation of $\mathbf{Z}$.
These indices are indirectly generated as the summation of initial indices and learnable offsets to provide stable initial status, where even-odd splitting indices are applied as initial indices.
Besides, to be compatible with different $T$, the offsets are learned in a normalized form, namely offset scales $\in (-1,1)$.
The offset scale $\mathbf{A}_S\in \mathbb{R}^{\frac{T}{2}}$ for relatively static group and $\mathbf{A}_D\in \mathbb{R}^{\frac{T}{2}}$ for dynamic group are learned as follows,
\begin{equation}
\label{Eq. offset scale}
\resizebox{0.9\hsize}{!}{$
\left\{\begin{aligned} 
\mathbf{A}_S &= \mathrm{Tanh}(\mathop{Vec}\limits_{dim=1,2}\left(\sigma(\frac{\mathbf{Z}\cdot (\mathbf{Z})^T}{\sqrt{d_T}})\right)^T\cdot  \mathbf{W}_S + \mathbf{b}_S) \\
\mathbf{A}_D &= \mathrm{Tanh}(\mathop{Vec}\limits_{dim=1,2}\left(\sigma(-\frac{\mathbf{Z}\cdot (\mathbf{Z})^T}{\sqrt{d_T}})\right)^T\cdot  \mathbf{W}_D + \mathbf{b}_D)    
\end{aligned}\right.,
$}
\end{equation}
where $\sigma$ denotes softmax function. 
$\mathbf{W}_S\in \mathbb{R}^{T^2 \times \frac{T}{2}}$, $\mathbf{W}_D\in \mathbb{R}^{T^2 \times \frac{T}{2}}$, $\mathbf{b}_S \in \mathbb{R}^{\frac{T}{2}}$, and $\mathbf{b}_D \in \mathbb{R}^{\frac{T}{2}}$ are all learnable embedding parameters.
Then the index vectors $\mathbf{I}_S\in \mathbb{R}^{\frac{T}{2}}$ and $\mathbf{I}_D\in \mathbb{R}^{\frac{T}{2}}$ can be denoted as,
\begin{equation}
\label{Eq. offset generation}
\left\{\begin{aligned} 
    \mathbf{I}_S[i] &= 2i + \mathbf{A}_S L  \\
    \mathbf{I}_D[i] &=  2i + 1 + \mathbf{A}_D L
\end{aligned}\right.\mathrm{for}\;\;i = 0,1,2,...,\frac{T}{2}-1,
\end{equation}
where hyperparameter $L$ presents the allowable range of offsets. 
Then we limit the range of $\mathbf{I}_S$ and $\mathbf{I}_D$ within $[0,T-1]$.

Based on obtained $\mathbf{I}_S$ and $\mathbf{I}_D$, frame features $\mathbf{X}$ are divided and interpolated into global context features $\mathbf{X}_S\in\mathbb{R}^{\frac{T}{2}\times H\times W\times C}$ and dynamic features $\mathbf{X}_D\in\mathbb{R}^{\frac{T}{2}\times H\times W\times C}$,
\begin{equation}
\label{Eq. split frames}
\resizebox{0.9\hsize}{!}{$
\left\{\begin{aligned}
\mathbf{X}_S[i] = \left(\mathbf{I}_S[i]-\left\lfloor\mathbf{I}_S[i]\right\rfloor\right)\mathbf{X}\left[\left\lceil\mathbf{I}_S[i]\right\rceil\right]+\left(\left\lceil\mathbf{I}_S[i]\right\rceil - \mathbf{I}_S[i]\right)\mathbf{X}\left[\left\lfloor\mathbf{I}_S[i]\right\rfloor\right]\\ 
\mathbf{X}_D[i] = \left(\mathbf{I}_F[i]-\left\lfloor\mathbf{I}_F[i]\right\rfloor\right)\mathbf{X}\left[\left\lceil\mathbf{I}_F[i]\right\rceil\right]+\left(\left\lceil\mathbf{I}_S[i]\right\rceil - \mathbf{I}_F[i]\right)\mathbf{X}\left[\left\lfloor\mathbf{I}_F[i]\right\rfloor\right]
\end{aligned}\right. ,
$}
\end{equation}
where $i = 0,1,2,...,\frac{T}{2}-1$. 
$\left\lfloor\cdot\right\rfloor$ and $\left\lceil\cdot\right\rceil$ denotes floor / ceil function that rounds down / up splitting indices respectively.
\subsection{Lifting-based Aggregation and Disentanglement}
Based on preliminary estimations $\mathbf{X}_S$ and $\mathbf{X}_D$, we propose Lifting-based Aggregation-Decoupling Module (LADM) to further disentanglement emotion-related dynamic features from global context via lifting scheme.
We additionally exploit mutual relation of $\mathbf{X}_S$ and $\mathbf{X}_D$ for better disentanglement, differing from vanilla lifting implementation.

LADM first utilizes the updater $\mathcal{U}$ to aggregate $\mathbf{X}_S$ and $\mathbf{X}_D$ to obtain refined global context estimation$\mathbf{Y}_S$.
The emotion-irrelevant global context remaining in preliminary estimated $\mathbf{X}_D$ is further squeezed out by the updater $\mathcal{U}$ and be absorbed into $\mathbf{Y}_S$.
Then, $\mathbf{Y}_S$ is exploited to extract irrelevant context features conditioned on $\mathbf{X}_D$ from itself and strip them from prior dynamic estimation $\mathbf{X}_D$, so as to purify the emotion-related dynamics $\mathbf{Y}_D$ by the predictor $\mathcal{P}$.
\begin{equation}
\label{Eq. Lifting}
\left\{\begin{aligned} 
    \mathbf{Y}_S &= \mathop{Vec}\limits_{dim=1,2,3}\mathbf{X}_S + \mathcal{U}(\mathbf{X}_D|\mathbf{X}_S)  \\
    \mathbf{Y}_D &= \mathop{Vec}\limits_{dim=1,2,3}\mathbf{X}_D - \mathcal{P}(\mathbf{Y}_S|\mathbf{X}_D)
\end{aligned}\right.,
\end{equation}
where $\mathcal{U}(\mathbf{X}_D|\mathbf{X}_S)$ implies extracting required information from $\mathbf{X}_D$ conditioned on $\mathbf{X}_S$, and $\mathcal{P}(\mathbf{Y}_S|\mathbf{X}_D)$ implies the reverse.
$\mathcal{U}$ and $\mathcal{P}$ adopt cross-attention mechanism to generate query, key, and value tensors as follows,
\begin{equation}
\resizebox{0.9\hsize}{!}{$
\left\{\begin{aligned} 
&\mathbf{X'}_S = \mathop{Vec}\limits_{dim=1,2,3}\mathbf{X}_S,\; \mathbf{X'}_D = \mathop{Vec}\limits_{dim=1,2,3}\mathbf{X}_D\\
&\mathbf{Q}_U = \frac{\mathbf{X'}_S\cdot \mathbf{W}_{QU}}{\sqrt{C}}, \mathbf{K}_U = \mathbf{X'}_D\cdot \mathbf{W}_{KU}, \mathbf{V}_U = \mathbf{X'}_D\cdot \mathbf{W}_{VU},\\ 
&\mathbf{Q}_P = \frac{\mathbf{X'}_D\cdot \mathbf{W}_{QP}}{\sqrt{C}}, \mathbf{K}_P = \mathbf{Y}_S\cdot \mathbf{W}_{KP}, \mathbf{V}_P = \mathbf{Y}_S\cdot \mathbf{W}_{VP}, 
\end{aligned}\right.
$}
\end{equation}
where $\mathbf{W}_{QU}$, $\mathbf{W}_{KU}$, $\mathbf{W}_{VU}$, $\mathbf{W}_{QP}$, $\mathbf{W}_{KP}$, and $\mathbf{W}_{VP}\in \mathbb{R}^{C\times C}$ are all learnable parameters of the same shape.
Then the updater $\mathcal{U}$ and predictor $\mathcal{P}$ are described by 
\begin{equation}
\label{Eq. Updata_and_Predict}
\left\{\begin{aligned} 
&\mathcal{U}(\mathbf{X}_D|\mathbf{X}_S) = \mathbf{MLP}\left(\sigma\left(\mathbf{Q}_U\cdot\mathbf{K}_U^T\right)\mathbf{V}_U\right)\\
&\mathcal{P}(\mathbf{Y}_S|\mathbf{X}_D) = \mathbf{MLP}\left(\sigma\left(\mathbf{Q}_P\cdot\mathbf{K}_P^T\right)\mathbf{V}_P\right)
\end{aligned}\right.,
\end{equation}
where $\mathbf{MLP}$ denotes stacked fully connected layer and GELU activation, $\sigma$ denotes softmax function.
The obtained $\mathbf{Y}_D$ in Eq. \ref{Eq. Lifting} is leveraged for facial expression recognition.
Besides, $\mathbf{Y}_S$ is further constraint to incorporate the global context information by disentanglement loss.

\subsection{Decoupling Loss\label{Sec. HybridLoss}}
For final expression recognition, dynamic features $\mathbf{Y}_D$ is flattened and projected to a classification vector $\boldsymbol{\kappa}\in \mathbb{R}^{N_C}$ by MLP, where $N_C$ denotes the number of emotion categories.
Disentanglement loss $\mathcal{L}$ is applied to $\boldsymbol{\kappa}$ and $\mathbf{Y}_S$ simultaneously, which is composed of task-specific loss $\mathcal{L}_{Task}$ and global context loss $\mathcal{L}_{Lift}$.
We adopt cross entropy loss as $\mathcal{L}_{Task}$ for DFER tasks, and thus rewrite $\mathcal{L}_{Task}$ as $\mathcal{L}_{CLS}$.
\begin{equation}
    \mathcal{L} = \mathcal{L}_{CLS} + \mathcal{L}_{Lift}.     
\end{equation}

Inspired by \cite{lifting-deep}, $\mathcal{L}_{Lift}$ imposes a restraint on $\mathbf{Y}_S$ to have the same local average with $\mathbf{X}$ to force $\mathbf{Y}^j_S$ incorporate all the global context information.
Local average is calculated along spatial dimension for $\{T,C\}$ dimensions of $\mathbf{X}$, since global context features should maintain spatial invariance across time.
Huber loss is adopted for $\mathcal{L}_{Lift}$.
Suppose there are $N$ video clips for one batch training and the label of $k$-th clip is $y_k$, then the components of $\mathcal{L}$ can be described by, 
\begin{equation}
\label{Eq. loss}
\resizebox{0.9\hsize}{!}{$
\left\{\begin{aligned} 
    &\mathcal{L}_{CLS} = -\sum_{k=1}^{N}\sum_{c=1}^{N_C}\mathbf{1}(c=y_k)log(\sigma(\boldsymbol{\kappa})[c])\\
    &\mathcal{L}_{Lift} = \sum_{k=1}^{N}\mathrm{Huber}\left(\mathop{Avg}\limits_{dim=T,C}(\mathbf{Y}_S - \mathrm{P}_{T\downarrow 2}(\mathbf{X}))\right)\\
\end{aligned}\right.,$}
\end{equation}
where $\mathbf{1}(\cdot)\in {0,1}$ is an indicator function equaling to 1 when the inner expression is true. 
$\mathop{Avg}$ denotes average calculation for specific dimensions $dim$.
$\mathrm{P}_{T\downarrow 2}(\cdot)$ represents temporal average pooling by a factor of 2.

\section{Experiments}
\subsection{Datasets and Metrics\label{Sec. Datasets}}
\noindent
\textbf{Datasets.}
We conduct evaluation on three important in-the-wild DFER datasets including DFEW\cite{DFEW2020} with 16,372 videos, FERV39k\cite{FERV39k} with 38,935 videos, and MAFW\cite{MAFW} with 10,045 videos.
Instead of lab-controlled videos, they collect videos from movies, TV dramas or other media sources.
They are challenging due to that limited frames relevant to labeled emotion are temporally diluted in a complex video scenario with occlusions and pose changes.
The category number of emotions is 7 for DFEW / FERV39k and 11 for MAFW, which all include neural emotion.
Note that three datasets all have long-tailed distribution issue for different emotions.
We follow the settings of these datasets for evaluation.
Specifically, DFEW and MAFW both provide 5-fold cross-validation settings, while FERV39k provides a train-test splitting setting.

\noindent
\textbf{Metrics.}
Consistent with prior researches\cite{RethinkM3DFEL,MAEDFER,IAL}, the unweighted average recall (UAR) and weighted average recall (WAR) as metrics. 
UAR is equivalent to the average accuracy of all expression categories ignoring the number of samples per class, while WAR represents the overall average accuracy of all samples.
We report the average metrics over three runs for FERV39k. As for DFER and MAFW with 5-fold cross-validation, we report metrics from a single run.

\subsection{Implementation Details\label{Sec. Implementation}}
\noindent
\textbf{Training Details.}
IFDD is implemented by PyTorch and trained on NVIDIA RTX 3090 for 100 epochs.
We utilize AdamW optimizer and cosine scheduler with 1e-4 initial learning rate and 1e-3 weight decay, where the former 10 epochs adopt warm-up strategy with 1e-6 learning rate.
Training sets from aforementioned datasets are further divided into training and validation set at a ratio of 4:1.

\noindent
\textbf{Data Preprocess.}
Fixed number $T_0$ of frames are uniformly sampled from videos and resized into the size of $H_0\times W_0$ as clips for training and inference.
$\{T_0, H_0 ,W_0\}$ are set to \{16,224,224\} for DFEW and FERV39k, and \{32,224,224\} for MAFW.
Data augmentation methods including random horizontal flip and random crop are adopted.

\begin{table}[t]
        \centering
        \setlength{\tabcolsep}{1mm}
        \small
        \begin{tabular}{ccccccc}
        \toprule
        \multicolumn{3}{c}{IFDD Variants} & \multicolumn{2}{c}{Metrics $(\%)$}\\
        \cmidrule(l){1-3}\cmidrule(l){4-5}
        Types & ISSM & LADM & UAR & WAR \\
        \midrule
        \multirow{4}{*}{2DCNN} & \XSolidBrush & \XSolidBrush &  47.68 & 59.93 \\
        & \Checkmark & \XSolidBrush & 52.06 & 65.40 \\
        & \XSolidBrush & \Checkmark &  54.24 & 68.05 \\
        & \Checkmark & \Checkmark & \textbf{56.72} & \textbf{70.01} \\
        \midrule
        \multirow{4}{*}{3DViT} & \XSolidBrush & \XSolidBrush &  57.00 & 68.60 \\
        & \Checkmark & \XSolidBrush & 55.53 & 71.68 \\
        & \XSolidBrush & \Checkmark &  58.67 & 72.44 \\
        & \Checkmark & \Checkmark & \textbf{61.19} & \textbf{73.82} \\
        \bottomrule
        \end{tabular}
        \caption{Analysis of ISSM and LADM on DFEW dataset.\label{Tab. OverviewAblation}}
\end{table}

\begin{table}[t]
    \center
    \small
    \setlength{\tabcolsep}{1mm}
    \renewcommand{\arraystretch}{1.1}
    \begin{tabular}{cccc}
    \toprule
    Splitting Dependency & Splitting Manner & UAR(\%) & WAR(\%) \\
    \hline
    Even-odd Assumption & Sampling & 58.67 & 72.44  \\
    Entire Features & Interpolation & 59.95 & 73.34 \\
    Temporal Correlation & Weighting & 58.53 & 72.68 \\
    \hline
    Temporal Correlation & Interpolation & \textbf{61.19} & \textbf{73.82}  \\
    \bottomrule
    \end{tabular}
    \caption{Evaluation on adaptive splitting methods with different settings. IFDD follows the settings at 4th line. 
    \label{Tab. Splitting}}
\end{table}

\begin{table}[t]
        \centering
        \renewcommand{\arraystretch}{1.2}
        \setlength{\tabcolsep}{1mm}
        \small
        \begin{tabular}{ccccc}
        \toprule
        \multirow{2}{*}{Mutual Relation} & \multirow{2}{*}{Network} & \multirow{2}{*}{Related Methods} & \multicolumn{2}{c}{Metrics (\%)}\\
        \cline{4-5}
         & & & UAR & WAR \\
        \hline
        w/o & CNN & DAWN, PIRNet & 60.45 & 72.19 \\
        w/o & Transformer & LGLFormer & 59.59 & 72.53 \\
        w/ & Transformer & LADM (Our) & \textbf{61.19} & \textbf{73.82} \\
        \bottomrule
        \end{tabular}
        \caption{Comparison on different Updater/Predictor designs.\label{Tab. LiftingScheme}}
\end{table}

\begin{table}[t]
        \centering
        \setlength{\tabcolsep}{1mm}
        \small
        \begin{tabular}{c|ccc}
        \toprule
        Global Context Loss & Constrained Dim. & UAR (\%) & WAR (\%)\\
        \midrule
        w/o & - & 59.58 & 72.87 \\
        w/ & $\{T,H,W,C\}$ & 59.74 & 73.17 \\
        w/ & $\{T,C\}$ & \textbf{61.19} & \textbf{73.82} \\
        \bottomrule
        \end{tabular}
        \caption{Evaluation on global context loss.\label{Tab. GlobalContextLoss}}
\end{table}

\noindent
\textbf{Network Settings.}
We explore two backbone types for IFDD, i.e., IFDD-2DCNN and IFDD-3DViT.
MobileNetV2 is adopted for IFDD-2DCNN, and MViT-S is adopted for IFDD-3DViT.
Detailed backbone settings for IFDD-2DCNN and IFDD-3DViT are illustrated in the extended version.
Besides, compressing factor $\frac{1}{D_c}$ of $Conv(\cdot)$ in Eq. \ref{Eq. TemporalToken} is set to 4 for IFDD-3DViT and 1 for IFDD-2DCNN respectively.
Channel number $d_T$ of temporal tokens $\textbf{Z}$ is set to 128 for the both.
Allowable Range $L$ of Eq. \ref{Eq. offset generation} is discussed in the extended version.

\begin{figure*}[!ht]
    \centering
    \includegraphics[width=0.95\textwidth]{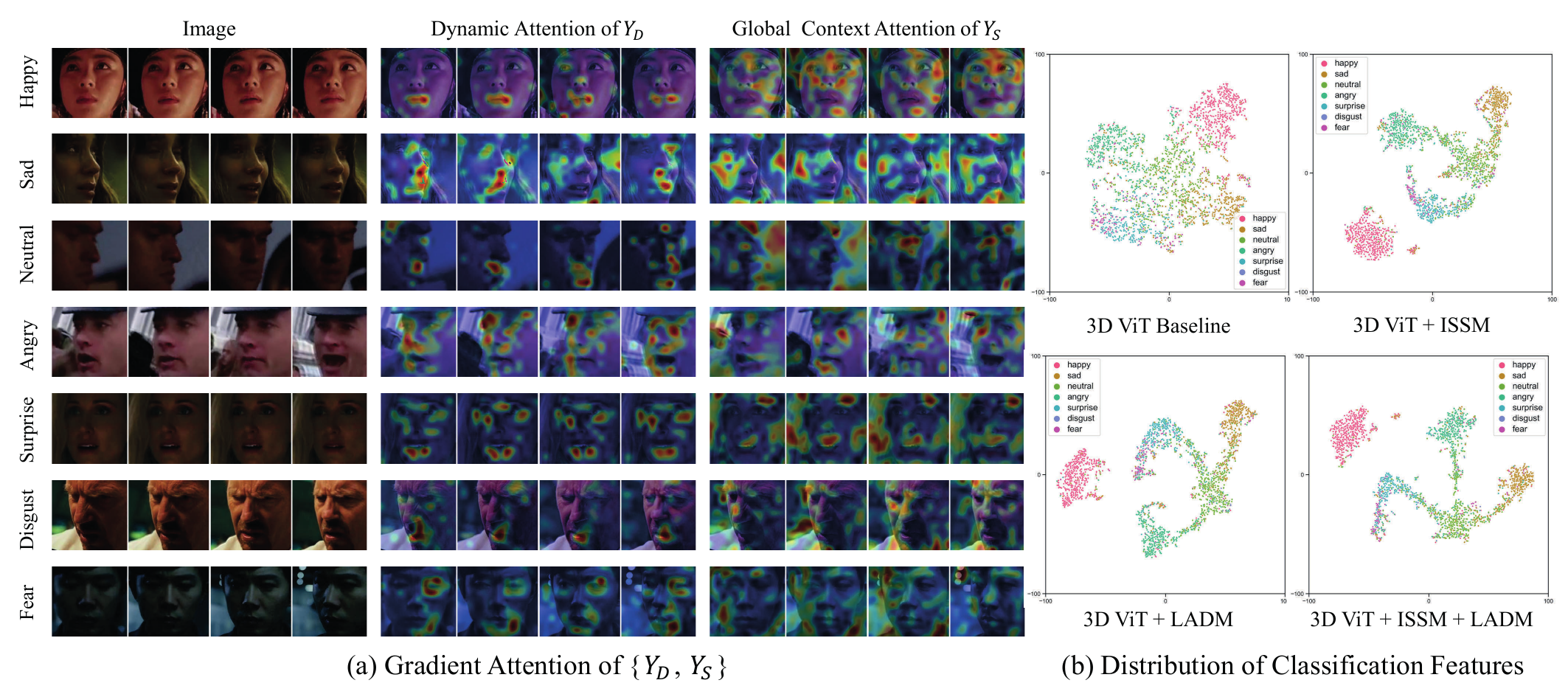}
    \caption{Visualization analysis on the gradient attention of $\{Y_D,Y_S\}$ by Grad-CAM in left subimage and the distribution of classification features by t-SNE in right subimage.
    For left subimage, clips and attention heatmaps are shown in different columns, while basic emotions are shown in different rows.
    Details can be found in ablation study and extended version.\label{fig:LiftFeature} 
    } 
\end{figure*}

\subsection{Ablation Study}
Following \cite{RethinkM3DFEL}, the experiments for ablation study are mainly conducted on DFEW dataset.
Due to limited space, we mainly present detailed ablation analysis for IFDD-3DViT.
Besides, additional ablation analysis can be found in extended version.

\noindent
\textbf{Overview of ISSM and LADM.}
We first investigate the effectiveness of ISSM and LADM in Table \ref{Tab. OverviewAblation}.
When solely using ISSM module, $\mathbf{X}_D$ is regarded as $\mathbf{Y}_D$.
As for solely using LADM, ISSM is substituted by even-odd splitting.
Compared to baselines, ISSM and LADM both show remarkable effectiveness in recognition accuracy.
Solely using LADM tend to gain higher WAR than ISSM, highlighting the second refinement stage of IFDD disentanglement process.

\noindent
\textbf{Adaptive Splitting Settings.}
We investigate the learning dependency and the splitting manner of ISSM in Table \ref{Tab. Splitting}, and additional details are illustrated in extended version.
Notably, the prior constraint of temporal correlation yields a $+0.48$ improvement in WAR, demonstrating the negative impacts of being exposed to the whole spatiotemporal features with excessive emotion-irrelevant global contents.
As for splitting manner, index interpolation manner outperforms weighting manner, which introduces stronger prior separation between static and dynamic frames.

\noindent
\textbf{Design of Learnable Lifting Scheme.}
Considering prior related methods, we evaluate the impacts of introducing mutual relation for LADM.
Detailed settings can be found in extended version.
Table \ref{Tab. LiftingScheme} show that the involvement of mutual relation contributes to $1.78\%$ improvement in WAR, demonstrating that lifting-based LADM profits from the interaction between global context and dynamic features.

\noindent
\textbf{Effectiveness of Global Context Loss.}
We investigate into whether to apply global context loss $\mathcal{L}_{Lift}$ and how tight we should impose constraints on $\textbf{Y}_S$ for different dimensions.
As shown in Table \ref{Tab. GlobalContextLoss}, the absence of $\mathcal{L}_{Lift}$ leads to a $2.63\%$ drop in UAR.
Besides, the constraint on local average for $\{T,C\}$ dimensions outperform $\{T,H,W,C\}$ dimensions.
It corroborates that $\textbf{Y}_S$ should be led to absorb static spatial information latent in temporal dimension.

\noindent
\textbf{Visualization Analysis.}
We further visualize the gradient attention of $Y_F$ and $Y_S$ extracted after the LayerNorm layer of LADM, and the distribution of classification features from the last linear layer. 
As revealed in Fig. \ref{fig:LiftFeature}(a), the disentangled dynamics $Y_D$ mainly focus on emotion-related features especially lips and eyes.
In contrast, $Y_S$ encompasses the entire faces including emotion-irrelevant background regions.
It can also be found that $Y_D$ focuses on different regions for different emotions, e.g., lips for \emph{Happy}, eyes for \emph{Sad}, and eyebrows for \emph{Surprise}.
Fig. \ref{fig:LiftFeature}(b) demonstrates that ISSM and LADM of IFDD considerably refine the discriminative and separable properties for embedding features.

\begin{table*}[htb]
        \center
        \small
        \setlength{\tabcolsep}{0.8mm}
        \renewcommand{\arraystretch}{1.1}
        \begin{tabular}{lcccccccccccc}
        \toprule
        \multirow{2}{*}{Methods} & \multicolumn{2}{c}{Efficiency} & \multicolumn{7}{c}{Accuracy of Each Emotion $(\%)$} & \multicolumn{2}{c}{Metrics $(\%)$}\\
        \cmidrule(l){2-3} \cmidrule(l){4-10} \cmidrule(l){11-12}
        & Params (M) & FLOPs (G)  & \emph{Happy} & \emph{Sad} & \emph{Neutral} & \emph{Angry} & \emph{Surprise} & \emph{Disgust} & \emph{Fear} & UAR & WAR 
        \\ 
        \midrule 
        C3D\cite{C3D} & 78  & 39 & 75.17 & 39.49 & 55,11 & 62.49 & 45.00 & 1.38 & 20.51 & 42.74 & 53.54 \\
        3D ResNet-18\cite{3DResNet}  & 33 & 8 & 76.32 & 50.21 & 64.18 & 62.85 & 47.52 & 0.00 & 24.56 & 46.52 & 58.27 \\
        R(2+1)D-18\cite{R21D} & 33  & 42 & 79.67 & 39.07 & 57.66 & 50.39 & 48.26 & 3.45 & 21.06 & 42.79 & 53.22 \\
        I3D\cite{I3D} & -  & 7 & 78.61 & 44.19 & 56.69 & 55.87 & 45.88 & 2.07 & 20.51 & 43.40 & 54.27 \\
        P3D\cite{P3D} & -  & - & 74.85 & 43.40 & 54.18 & 60.42 & 50.99 & 0.69 & 23.28 & 43.97 & 54.47 \\
        ResNet18+GRU\cite{Former-DETR} & -  & 8 & 82.87 & 63.83 & 65.06 & 68.51 & 52.00 & 0.86 & 30.14 & 51.68 & 64.02 \\
        ResNet18+LSTM\cite{Former-DETR} & -  & 8 & 83.56 & 61.56 & 68.27 & 65.29 & 51.26 & 0.00 & 29.34 & 51.32 & 63.85 \\
        Former-DFER\cite{Former-DETR} & 18 & 9 & 84.05 & 62.57 & 67.52 & 70.03 & 56.43 & 3.45 & 31.78 & 53.69 & 65.70 \\
        STT\cite{STT} & - & - & 87.36 & 67.90 & 64.97 & 71.24 & 53.10 & 3.49 & 34.04 & 54.58 & 66.65 \\
        DPCNet\cite{DPCNet} & 51  & 10 & - & - & - & - & - & - & - & 55.02 & 66.32 \\
        NR-DFERNet\cite{NR-DFERNet} & -  & 6 & 88.47 & 64.84 & 70.03 & 75.09 & 61.60 & 0.00 & 19.43 & 54.21 & 68.19 \\
        GCA+IAL\cite{IAL} & 19  & 10 & 87.95 & 67.21 & 70.10 & 76.06 & 62.22 & 0.00 & 26.44 & 55.71 & 69.24\\
        Freq-HD\cite{FreqHD} & -  & - & 53.56 & 38.10 & 23.20 & \textbf{77.91} & 6.9 & \textbf{45.38} & \textbf{64.61} & 46.85 & 55.68\\
        M3DFEL\cite{RethinkM3DFEL} & 13.86  & \textbf{1.66} & 89.59 & 68.38 & 67.88 & 74.24 & 59.69 & 0.00 & 31.64 & 56.10 & 69.25 \\
        CDGT\cite{CDGT} & 16.46  & 8.34 & 89.90 & 70.86 & 63.99 & 73.77 & 58.88 & 17.24 & 39.47 & 59.16 & 70.07\\
        \hline
        MobileNetV2 (Baseline) & 1.55 & 3.73 & 84.46 & 52.24 & 62.17 & 59.54 & 57.14 & 0.00 & 18.23 & 47.68 & 59.93 \\
        MViT\cite{MViT2021} (Baseline) & 23.41 & 32.83 & 88.14 & 59.37 & 72.66 & 70.34 & 66.67 & 10.34 & 31.49 & 57.00 & 68.60 \\
        IFDD-2DCNN (ours) & \textbf{1.18}  & 3.86 & 90.80 & 71.24 & 69.48 & 70.80 & 65.99 & 0.00 & 28.73 & 56.72 & 70.01 \\
        IFDD-3DViT (ours) & 37.44 & 36.01 & \textbf{94.07} & \textbf{75.99} & \textbf{74.16} & 74.02 & \textbf{69.39} & 10.34 & 30.39 & \textbf{61.19} & \textbf{73.82} \\
        \bottomrule
        \end{tabular}
        \caption{Comparisons with the state-of-the-art methods on DFEW dataset. - means not reported in the corresponding paper.\label{Tab. DFEW}}
\end{table*}

\begin{table}[t]
    \centering
    \small
    \begin{tabular}{lcc}
    \toprule
    \multirow{2}{*}{Methods} & \multicolumn{2}{c}{Metrics $(\%)$} \\
    \cmidrule{2-3}
    & UAR & WAR \\
    \midrule
    C3D\cite{C3D} & 22.68 & 31.69\\
    3D ResNet-18\cite{3DResNet} & 26.67 & 37.57 \\
    R(2+1)D-18\cite{R21D} & 31.55 & 41.28 \\
    I3D\cite{I3D} & 30.17 & 38.78 \\
    ResNet18+LSTM\cite{FERV39k} & 30.92 & 42.59 \\
    Two VGG13+LSTM\cite{FERV39k} & 32.79 & 44.54 \\
    Former-DFER\cite{Former-DETR} & 37.20 & 46.85 \\
    STT\cite{STT} & 37.76 & 46.85 \\
    NR-DFERNet\cite{NR-DFERNet} & 33.99 & 45.97 \\
    GCA+IAL\cite{IAL} & 35.82 & 48.54 \\
    Freq-HD\cite{FreqHD} & 33.07 & 45.26 \\
    M3DFEL\cite{RethinkM3DFEL} & 35.94 & 47.67 \\
    IFDD-2DCNN (ours) & 36.53 & 47.91 \\
    IFDD-3DViT (ours) & \textbf{39.15} & \textbf{51.09} \\
    \bottomrule
    \end{tabular}
    \caption{Accuracy comparison on FERV39k dataset. \label{Tab. FERV39k}
    }
\end{table} 

\begin{table}[t]
    \centering
    \small
    \setlength{\tabcolsep}{3.2mm}
    \begin{tabular}{lcc}
    \toprule
    \multirow{2}{*}{Methods} & \multicolumn{2}{c}{Metrics $(\%)$} \\
    \cmidrule{2-3}
    & UAR & WAR \\
    \midrule
    3D ResNet-18\cite{3DResNet} & 25.58 & 36.65 \\
    ViT\cite{ViT} & 32.36 & 45.04 \\
    C3D\cite{C3D} & 31.17 & 42.25 \\
    ResNet18+LSTM\cite{MAFW} & 28.08 & 39.38 \\
    ViT+LSTM\cite{MAFW} & 32.67 & 45.56 \\
    C3D+LSTM\cite{MAFW} & 31.17 & 42.25 \\
    T-ESFL\cite{MAFW} & 33.28 & 48.18 \\
    Former-DFER\cite{Former-DETR} & 31.16 & 43.27 \\
    IFDD-2DCNN (ours) & 33.68 & 48.93 \\
    IFDD-3DViT (ours) & \textbf{39.31} & \textbf{53.92} \\
    \bottomrule
    \end{tabular}
    \caption{Accuracy comparison on MAFW dataset.\label{Tab. MAFW}
    }
\end{table}

\subsection{Comparison with State-of-the-Art Methods}
We evaluate IFDD and conduct comparisons with state-of-the-art methods on DFEW, FERV39k, and MAFW datasets. 
Involved previous methods are split into two categories, namely supervised methods and self-supervised methods.
Notably, for fair comparison, we only compare with supervised models, excluding those self-supervised methods pretrained on large-scale external sources.
The discussions with self-supervised methods are in extended version.

\noindent
\textbf{DFEW.}
As shown in Table \ref{Tab. DFEW}, we evaluate IFDD variants and compare with state-of-the-art methods under 5-fold cross-validation setting.
IFDD-3DViT and IFDD-2DCNN outperform previous supervised methods in both overall and per-emotion accuracy metrics.
For overall weighted accuracy WAR, IFDD-3DViT has taken a lead of $6.60\%$ percent over the best recorded supervised method M3DFEL\cite{RethinkM3DFEL} with $73.82\%$ WAR.
For per-class accuracy, IFDD-3DViT has shown efficacious recognition for \emph{Happy}, \emph{Sad}, and \emph{Neutral} emotions.
Besides, CNN and ViT baselines equipped with IFDD have incurred $3.49\%$ and $9.69\%$ increments in extra FLOPs, which is quite tolerable.
IFDD-2DCNN achieves considerable efficiency with few parameters and low computational cost, while still surpasses M3DFEL with $70.01$ WAR and $56.72$ UAR.

\noindent
\textbf{FERV39k.}
Accuracy comparison with prior methods on FERV39k dataset is shown in Table \ref{Tab. FERV39k}.
IFDD-3DViT outperforms the second-best supervised method M3DFEL\cite{RethinkM3DFEL} by $7.17\%$ in WAR and $8.93\%$ in UAR.
Besides, compared with temporal 2DCNNs, i.e., ResNet18+LSTM and ResNet+GRU, IFDD-2DCNN surpasses ResNet18+LSTM by $12.49\%$/$18.14\%$ in WAR/UAR and $51.75\%$ in FLOPs, exhibiting superior efficiency.

\noindent
\textbf{MAFW.}
Experimental results are shown in Table \ref{Tab. MAFW}.
IFDD-3DViT and IFDD-2DCNN both significantly outperform the second-best supervised method T-ESFL, especially IFDD-3DViT exceeding by 11.91\% in WAR and 18.12\% in UAR.

\section{Conclusion}
In this paper, to focus on emotion-related expression dynamics that are temporally diluted by irrelevant non-expression context in in-the-wild DFER, we propose IFDD framework to disentangle emotion-related dynamic features from global context in an implicit manner.
IFDD expands the framework of lifting scheme, and proposes ISSM and LADM for two-stage disentanglement.
Extensive experiments on in-the-wild DFER datasets demonstrate the superiority of IFDD over state-of-the-art supervised methods, as well as its compatibility with different backbones. 
Future directions include exploring IFDD framework for other dynamic-sensitive tasks, such as dynamic micro-expression recognition, optical flow estimation, and video compression.

\section*{Acknowledgements}
This work was supported by the National Natural Science Foundation of China (grant U2441244), and by Zhejiang Provincial Natural Science Foundation of China (grant LZ24F030006).

\bibliography{reference_AAAI}

\clearpage
\appendixpage
\section{A. Additional Implementation Details}
\subsection{Pseudocode for IFDD framework}
We intuitively represent the algorithm of IFDD via pseudocode, as shown in Algorithm \ref{alg. IFDD}.

\subsection{Backbone Details}
Here we illustrate the detailed process to extract multiscale features for the backbone of IFDD.
Recall that we have explored two types of networks for the backbone, namely MobileNetV2 and MViT-S, and thus obtain IFDD family including IFDD-2DCNN and IFDD-3DViT respectively.

Given video clip $\mathbf{F}\in\mathbb{R}^{T_0\times H_0\times W_0\times 3}$, the backbone first adopts an input convolution layer to downsample the temporal dimension by a factor of 2 and the spatial dimension by a factor of 4.
Recall that the spatial size $H_0\times W_0$ of input clips is $224\times 224$ for all datasets, after this downsampling, the spatial size of features for latter layers is $56\time 56$.
Then the backbone extracts multiscale features from the clip, i.e., a multiscale pyramid of features with $S$ stages.
Note that the temporal sizes of features at different stages are the same, and the spatial size are different.
In this way, the features at $j$-th stage are defined as $\mathbf{F}^j\in\mathbb{R}^{\frac{T_0}{2}\times \frac{H_0}{2^{j+1}}\times \frac{W_0}{2^{j+1}}\times C_{j}}$, where $j\in \{1,2,..,S\}$ and $C_{j}$ denotes the channel dimension.

For IFDD-2DCNN, only the former 5 bottlenecks of MobileNetV2 are utilized.
The 3-stage features with the spatial widths $\{56,28,14\}$ are extracted from 3-th, 4-th, and 5-th bottlenecks respectively.
The spatial sizes of extracted multiscale features by IFDD-3DViT are the same with IFDD-2DCNN.
Only the former 3 stages of MViT-S are used for the feature extraction.
3-stage features with the spatial widths $\{56,28,14\}$ are extracted from 1-th, 2-th, and 3-th stages of MViT-S respectively.
We remove 6 blocks in 3-th stage of MViT-S for further efficiency.
Since we leverage comprehensive features of MViT-S for fine-grained disentanglement, we also remove the class token of MViT-S.

The subsequent pyramid aggregation consists of two steps, namely compression and aggregation.
The compression progress applies dilated $3\times 3$ convolution layers with $\textrm{stride}=S-j+1$ to features $\mathbf{F}^j$ of different stage $j$ respectively, resulting in compressed features of the same spatiotemporal size.
Then aggregation progress concatenated these features along channel dimension, and compress them at channel dimension by a $1\times 1$ convolution layer, resulting in $\mathbf{X} \in\mathbb{R}^{\frac{T_0}{2}\times \frac{H_0}{2^{S+1}}\times \frac{W_0}{2^{S+1}}\times C_{S}}$.
$S$ is set to 3 in this paper.

\section{B. Additional Information for Ablation Study}
\subsection{Adaptive Splitting Settings}
We evaluate two extra learnable splitting methods of different settings to investigate the effectiveness of the learning dependency and the splitting manner of ISSM, namely $\mathrm{ISSM}^{\dag}$ and $\mathrm{ISSM}^{\ddag}$ as in Table \ref{Tab. Splitting_all}.

\begin{algorithm}[tb]
    \caption{Pseudocode for IFDD framework\label{alg. IFDD}}
    \textbf{Input}:  A video clip $\mathbf{F}$.\\
    \textbf{Parameter}: Network parameters $\theta$; $\rm{Backbone}_\theta$; $\rm{ISSM}_\theta$; $\mathcal{U}_\theta$ and $\mathcal{P}_\theta$ in LADM; Learning rate $lr$.\\
    \textbf{Output}: Recognition result.
    \begin{algorithmic}[1] 
    \STATE Extract latents feature $\mathbf{X}=\rm{Backbone}_\theta(\mathbf{F})$;
    \STATE Generate content-aware indices $\mathbf{I}_S, \mathbf{I}_D = \rm{ISSM}_\theta(\mathbf{X})$ by Eq. \ref{Eq. TemporalToken}, \ref{Eq. offset scale} and \ref{Eq. offset generation};
    \STATE Split $\mathbf{X}$ into to static and dynamic groups $\mathbf{X}_S, \mathbf{X}_D = \rm{Interpolate}(\mathbf{X}, \mathbf{I}_S, \mathbf{I}_D)$ by Eq. \ref{Eq. split frames};
    \STATE Aggregate $\mathbf{X}_S$ and $\mathbf{X}_D$ to get global context features: $\mathbf{Y}_S=\mathop{Vec}\limits_{dim=1,2,3}\mathbf{X}_S + \mathcal{U}(\mathbf{X}_D|\mathbf{X}_S)$;
    \STATE Purify dynamic features $\mathbf{Y}_D$ by filtering out global context from $\mathbf{X}_D$: $\mathbf{Y}_D = \mathop{Vec}\limits_{dim=1,2,3}\mathbf{X}_D - \mathcal{P}(\mathbf{Y}_S|\mathbf{X}_D)$;
    \STATE Project $\mathbf{Y}_D$ to classification vector $\boldsymbol{\kappa}$, and apply losses $\mathcal{L}_{CLS}$ and $\mathcal{L}_{Lift}$ in Eq. \ref{Eq. loss} to $\boldsymbol{\kappa}$ and $\mathbf{Y}_S$ respectively;
    \STATE Take gradient step on $\nabla_\theta(\mathcal{L}_{CLS}+\mathcal{L}_{Lift})$;
    \STATE Update $\theta=\theta+lr*\nabla_\theta$ by optimizer;
    \STATE \textbf{return} Recognition result $=\arg\max(\rm{softmax}(\boldsymbol{\kappa}))$.
    \end{algorithmic}
\end{algorithm}

\begin{table}[t]
    \center
    \renewcommand{\arraystretch}{1.2}
    \resizebox{\linewidth}{!}{
    \begin{tabular}{cccc}
    \toprule
    Methods & Splitting Dependency & Splitting Manner\\
    \hline
    Even-odd splitting & Prior Assumption & Sampling \\
    \hline
    $\textrm{ISSM}^\dag$ & Entire Features & Interpolation\\
    $\textrm{ISSM}^\ddag$  & Temporal Correlation & Weighting\\
    ISSM & Temporal Correlation & Interpolation \\
    \bottomrule
    \end{tabular}}
    \caption{Adaptive splitting methods with different settings. $\textrm{ISSM}^\dag$ learns the splitting index directly from entire spatiotemporal features, while $\textrm{ISSM}^\ddag$ utilize weighted scoring instead of interpolation for splitting.\label{Tab. Splitting_all}}
\end{table}

\begin{figure}[t]
    \centering
    \includegraphics[width=0.45\textwidth]{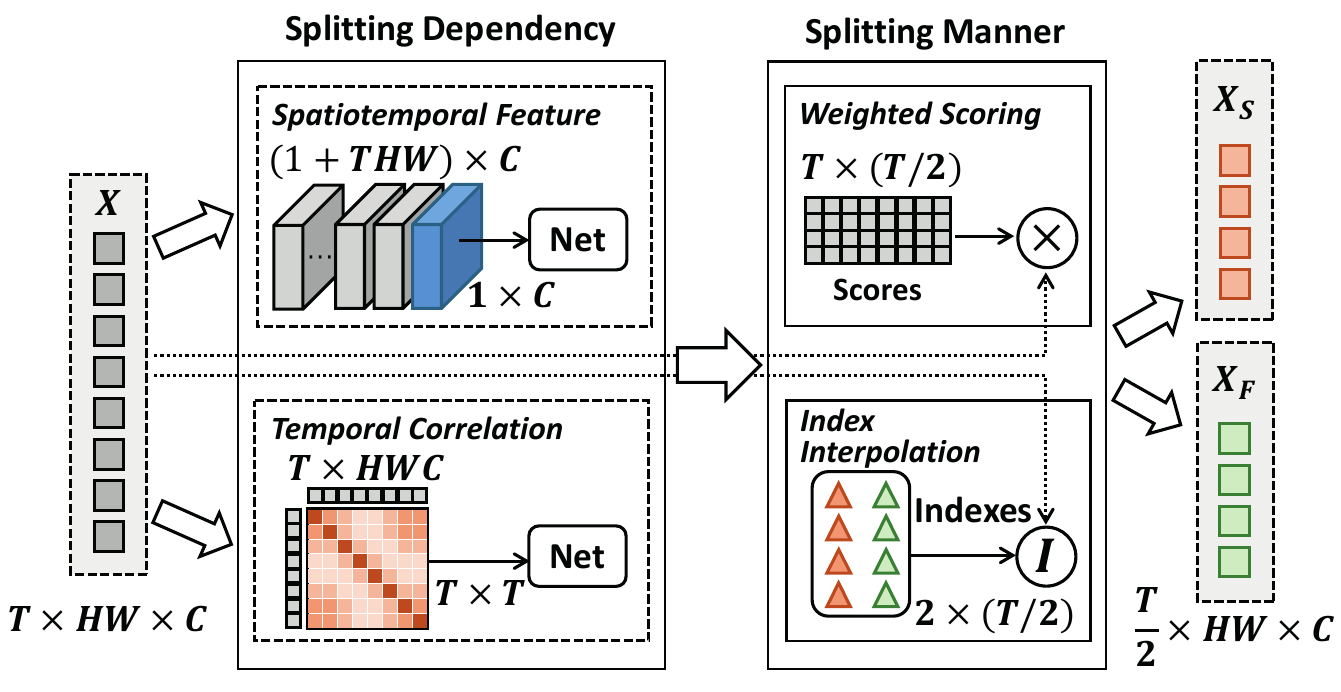}
    \caption{Schematic diagram of different ISSM variants. 
    T, H, and W are the temporal and spatial size of the input feature $\textbf{x}$ respectively. $I$ denotes linear interpolation and $\otimes$ denotes element-wise multiplication.\label{fig:Split_Variants}}
\end{figure}

\begin{table}[t]    
    \centering
    \renewcommand{\arraystretch}{1.5}
    \setlength\tabcolsep{3pt}
    \small
    \begin{tabular}{c|ccc}
    \toprule
    Initial Index & Allowable Range & UAR (\%) & WAR (\%)\\
    \midrule
    \multirow{2}{*}{Even-odd indices}
    & $\pm\frac{T}{4}$ & 59.82 & 73.24 \\
    & $\pm\frac{T}{2}$ & \textbf{61.19} & \textbf{73.82} \\
    \midrule
    \multirow{1}{*}{Midpoint ($\frac{T}{2}$)}
    & $\pm\frac{T}{2}$ & 60.27 & 73.67 \\
    \bottomrule
    \end{tabular}
    \caption{Evaluation on allowable range of offsets in ISSM. $T$ denotes the dimension of the temporal channel, specifically $T=8$ for IFDD implementation.\label{Tab. OffsetRange}}
\end{table}

\begin{table}[t]
    \centering
    \renewcommand{\arraystretch}{1.2}
    \setlength\tabcolsep{8pt}
    \small
    \begin{tabular}{cccc}
    \toprule
    \multicolumn{2}{c}{Different Orders} & \multicolumn{2}{c}{Metrics (\%)}\\
    \cmidrule(l){1-2}\cmidrule(l){3-4}
    The Former & The Latter & UAR & WAR\\
    \midrule
    Predictor & Updater & 59.29 & 72.88 \\
    Updater & Predictor & \textbf{61.19} & \textbf{73.82} \\
    \bottomrule
    \end{tabular}
    \caption{Evaluation on the order of aggregation and disentanglement.\label{Tab. UPOrder}}
\end{table}

Here we illustrate their details, as shown in Figure \ref{fig:Split_Variants}.
Suppose the temporal size of input feature $\textbf{X}$ is $T$,  $\mathrm{ISSM}^{\dag}$ utilize self-attention\cite{transformer} across the whole spatiotemporal dimensions to learn the splitting indices.
Specifically, a global token with the size $1\times C$ is concatenated with $\textbf{X}$ to get $\textbf{X}'\in \mathbb{R}^{(1+THW)\times C}$.
After self-attention, the global token is separated and compressed by an MLP layer to obtain two index vectors of the length $T/2$.

In contrast, $\mathrm{ISSM}^{\ddag}$ follows ISSM to learn the latent splitting features based on temporal correlation.
Differing from ISSM, $\mathrm{ISSM}^{\ddag}$ then projects the latent features to two $T\times \frac{T}{2}$ matrices, and use these matrices to weight $\textbf{X}$ at temporal dimension, resulting in two index vectors.

Other settings of $\mathrm{ISSM}^{\dag}$ and $\mathrm{ISSM}^{\ddag}$ follow ISSM.

\subsection{Design of Learnable Lifting Scheme}
Recall that, to evaluate the impacts of the introduced mutual relation for lifting scheme, we compared LADM with two variants.
Considering prior related methods, we also evaluate two types of networks to implement lifting scheme. 
The main differences of these variants lie in network type and whether involving mutual relation between emotion-related dynamics and irrelevant global context.

The variant at the first row of Table 3 in the main paper are designed based on 3D-CNN framework, similar to the framework of \cite{lifting-deep,PIRNet, WINNet}.
Two cascaded 3D convolution layers with 3D batch normalization and ReLU activation are applied for both predictor and updater.
The second-row variant in Table 3 follows the settings of LADM, apart from that updater and predictor in Eq. 5 of the main paper are replaced to $\mathcal{U}(\mathbf{X}_D|\mathbf{X}_D)$ and $\mathcal{P}(\mathbf{X}_S|\mathbf{X}_S)$ respectively.
The updater and predictor of these variants are all input single-branch features without the corresponding mutual relation.

\section{C. Additional Experiments for Ablation Study}

\begin{table*}[t]
    \centering    
    \begin{tabular}{cccccccccccc}
    \toprule
    \multicolumn{3}{c}{IFDD Variants} & \multicolumn{7}{c}{Accuracy of Each Emotion $(\%)$} & \multicolumn{2}{c}{Metrics $(\%)$}\\
    \cmidrule(l){1-3}\cmidrule(l){4-10}\cmidrule(l){11-12}
    Backbones & ISSM & LADM & \emph{Happy} & \emph{Sad} & \emph{Neutral} & \emph{Angry} & \emph{Surprise} & \emph{Disgust} & \emph{Fear} &UAR & WAR \\
    \midrule
    \multirow{4}{*}{IFDD-2DCNN} & \XSolidBrush & \XSolidBrush &  84.46 & 52.24 & 62.17 & 59.54 & 57.14 & 0.00 & 18.23 & 47.68 & 59.93 \\
    & \Checkmark & \XSolidBrush & 87.73 & 60.16 & 70.22 & 68.28 & 53.74 & 0.00& 24.31 & 52.06 & 65.40 \\
    & \XSolidBrush & \Checkmark &  89.57 & 64.38 & 70.79 & 73.10 & 59.18 & 0.00 & 22.65 & 54.24 & 68.05 \\
    & \Checkmark & \Checkmark &  90.80 & 71.24 & 69.48 & 70.80 & 65.99 & 0.00 & 28.73 & \textbf{56.72} & \textbf{70.01} \\
    \midrule
    \multirow{4}{*}{IFDD-3DViT} & \XSolidBrush & \XSolidBrush &  88.14 & 59.37 & 72.66 & 70.34 & 66.67 & 10.34 & 31.49 & 57.00 & 68.60 \\
    & \Checkmark & \XSolidBrush & 95.50 & 62.53 & 80.90 & 75.63 & 69.73 & 0.00 & 4.42 & 55.53 & 71.68 \\
    & \XSolidBrush & \Checkmark &  94.07 & 69.92 & 73.78 & 77.47 & 59.52 & 0.00 & 35.91 & 58.67 & 72.44 \\
    & \Checkmark & \Checkmark &  94.07 & 75.99 & 74.16 & 74.02 & 69.39 & 10.34 & 30.39 & \textbf{61.19} & \textbf{73.82} \\
    \bottomrule
    \end{tabular}
    \caption{Per-emotion Accuracy Comparison with baselines on DFEW dataset.\label{Tab. OverviewAblation}}
\end{table*}

\begin{figure}[htb]
    \centering    
    \captionsetup[subfloat]{labelformat=empty,position=above}
    \subfloat[\small{Compared with Prior Methods\quad\; Compared with Baseline}]{
            \rotatebox{90}{\small{\textrm{\qquad\qquad\qquad IFDD-3DViT}}}
            \includegraphics[width=0.45\textwidth]{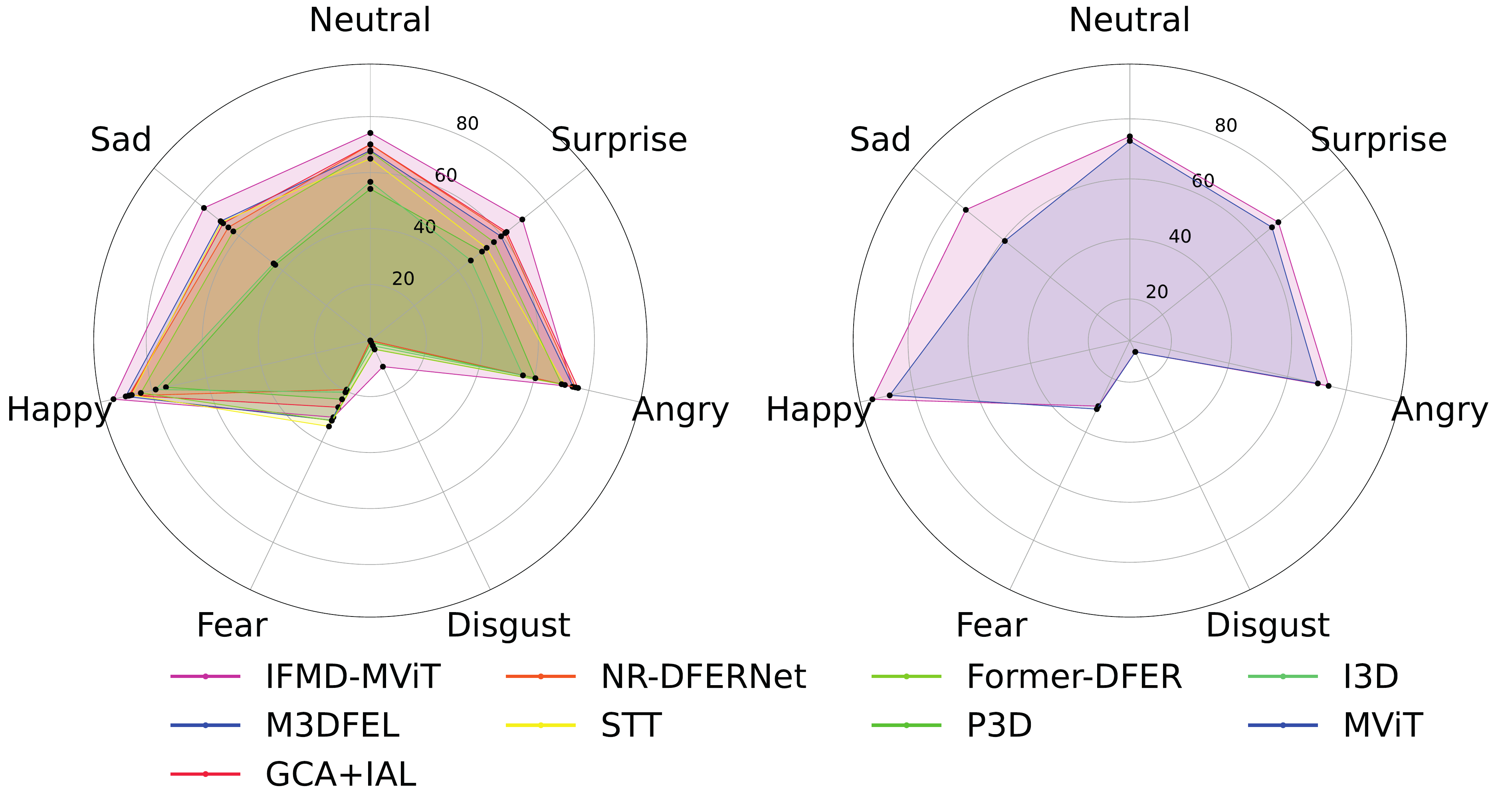}
    }
    \vspace{0.5mm}
    \subfloat{
            \rotatebox{90}{\small{\textrm{\qquad\qquad\quad IFDD-2DCNN}}}
            \includegraphics[width=0.45\textwidth]{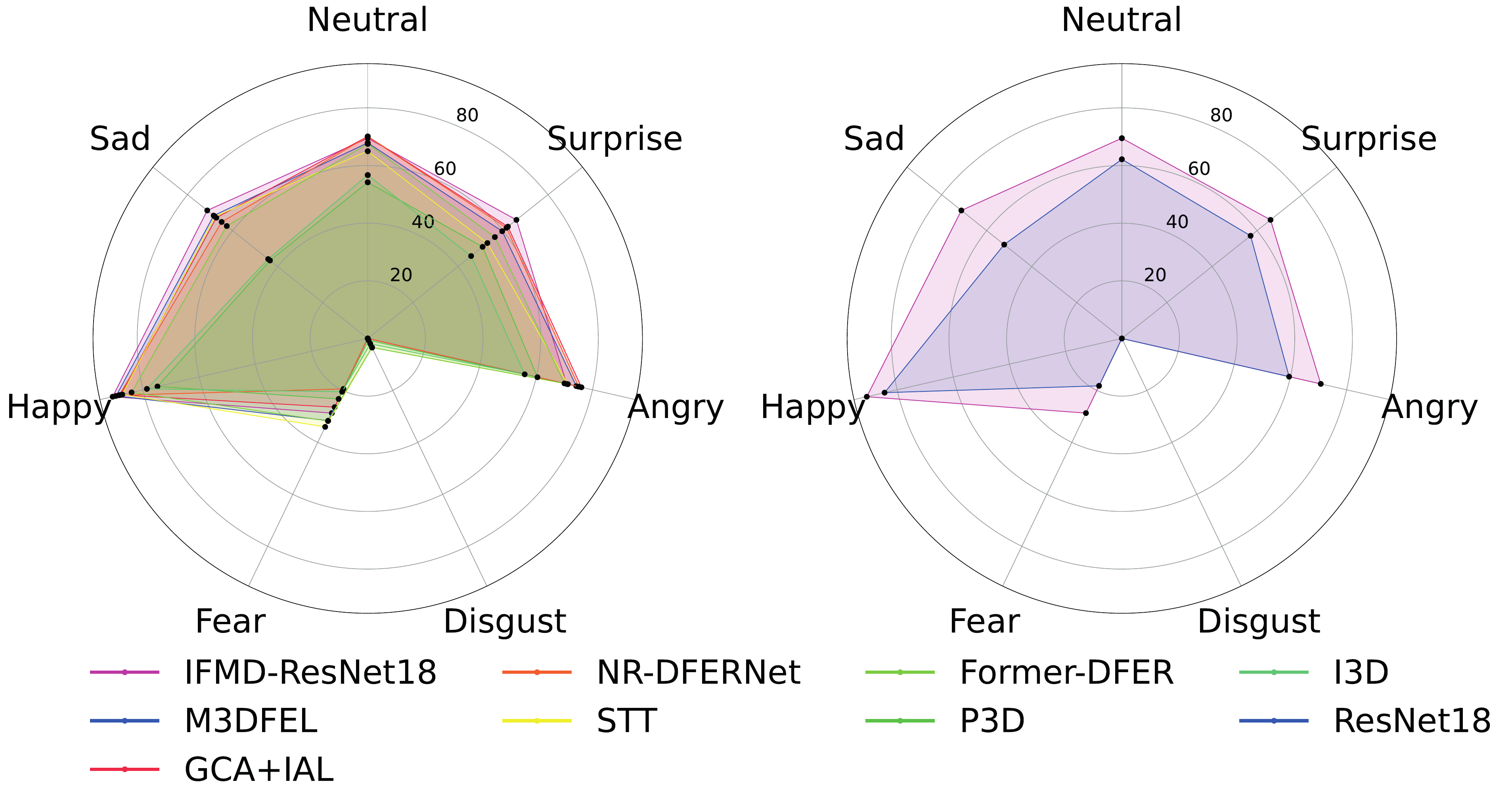}
    }
    \caption{Visual Comparison with baselines and prior methods for per-class accuracy on DFEW test set. The envelope area of a radar chart represents its corresponding average class accuracy, i.e., UAR.\label{fig:Radar} }
\end{figure}

\begin{figure}[!htb]
    \centering
    \captionsetup[subfloat]{labelformat=empty}
    \subfloat[2D-CNN Baseline]{
            \includegraphics[width=0.20\textwidth]{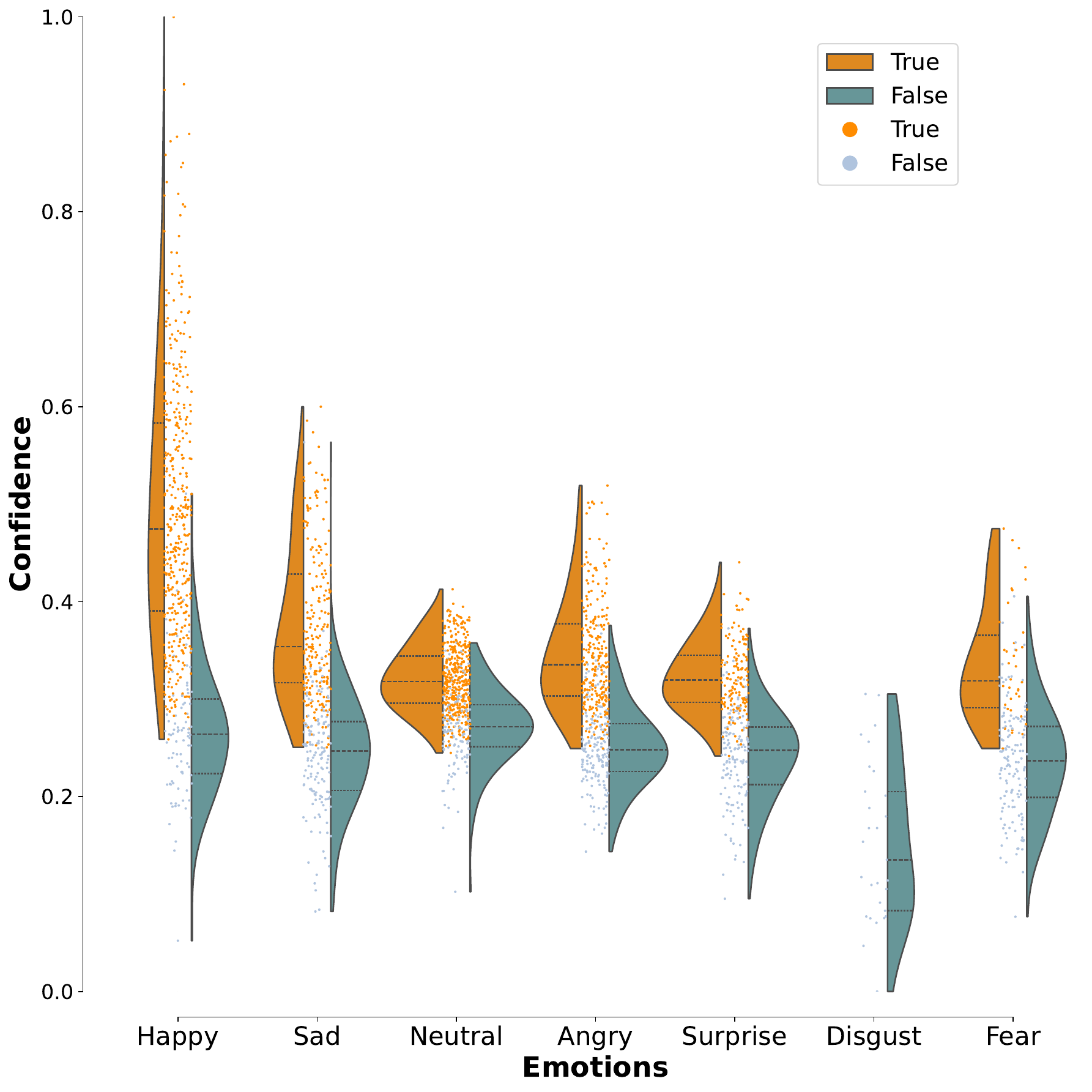}
    }
    \subfloat[2D-CNN Baseline with IFDD]{
            \includegraphics[width=0.20\textwidth]{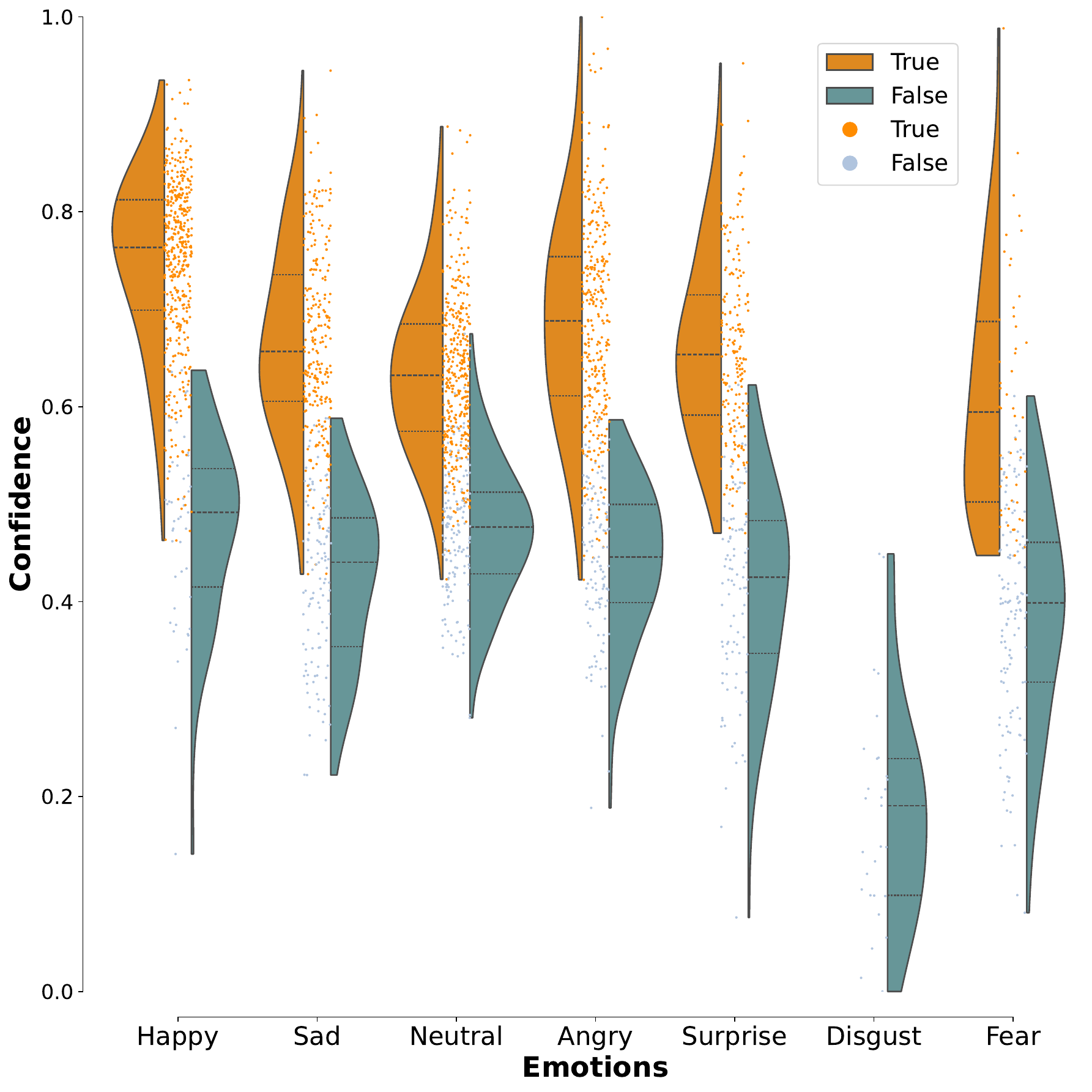}
    }
    \vspace{0.5mm}
    \subfloat[3D-ViT Baseline]{
            \includegraphics[width=0.20\textwidth]{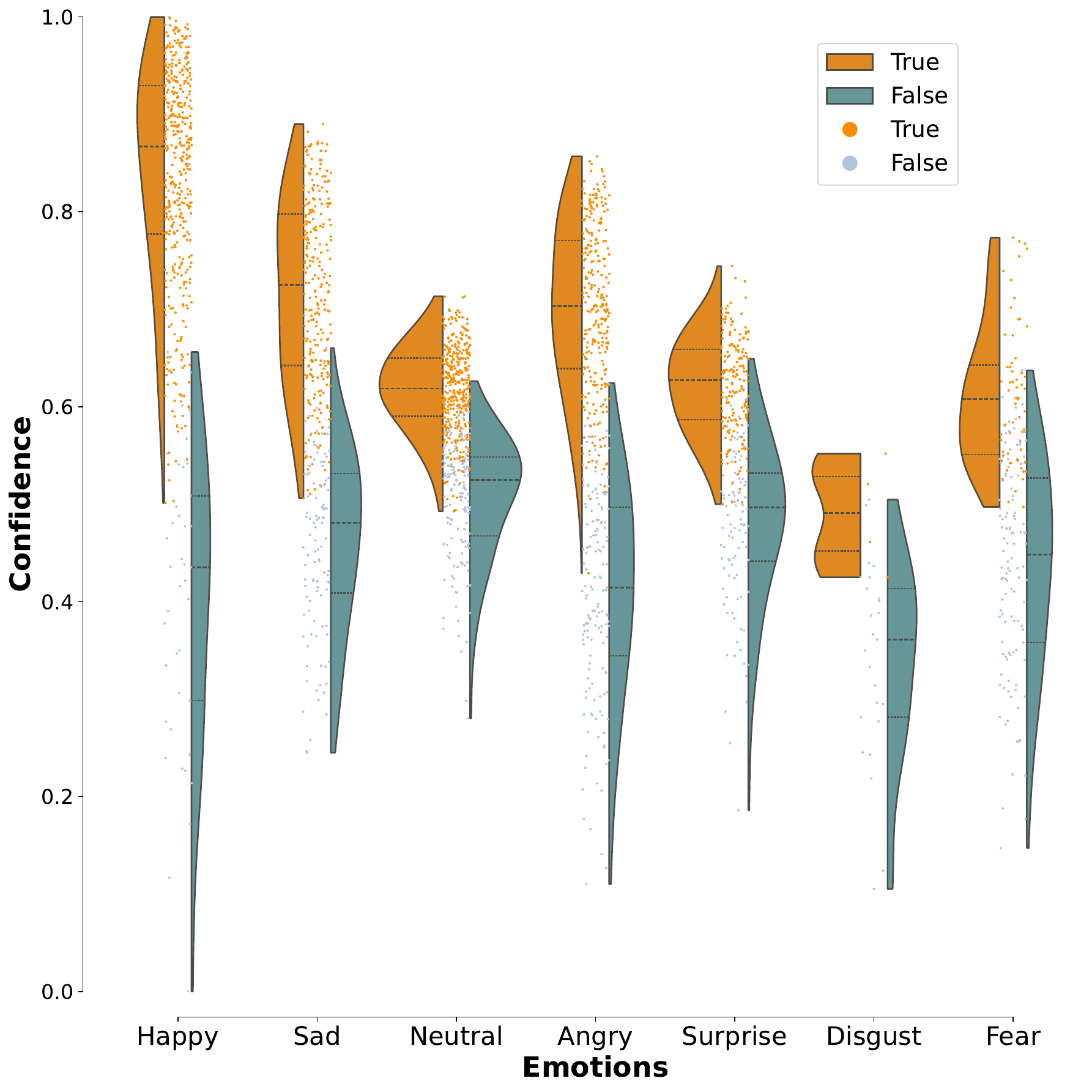}
    }
    \subfloat[3D-ViT Baseline with IFDD]{
            \includegraphics[width=0.20\textwidth]{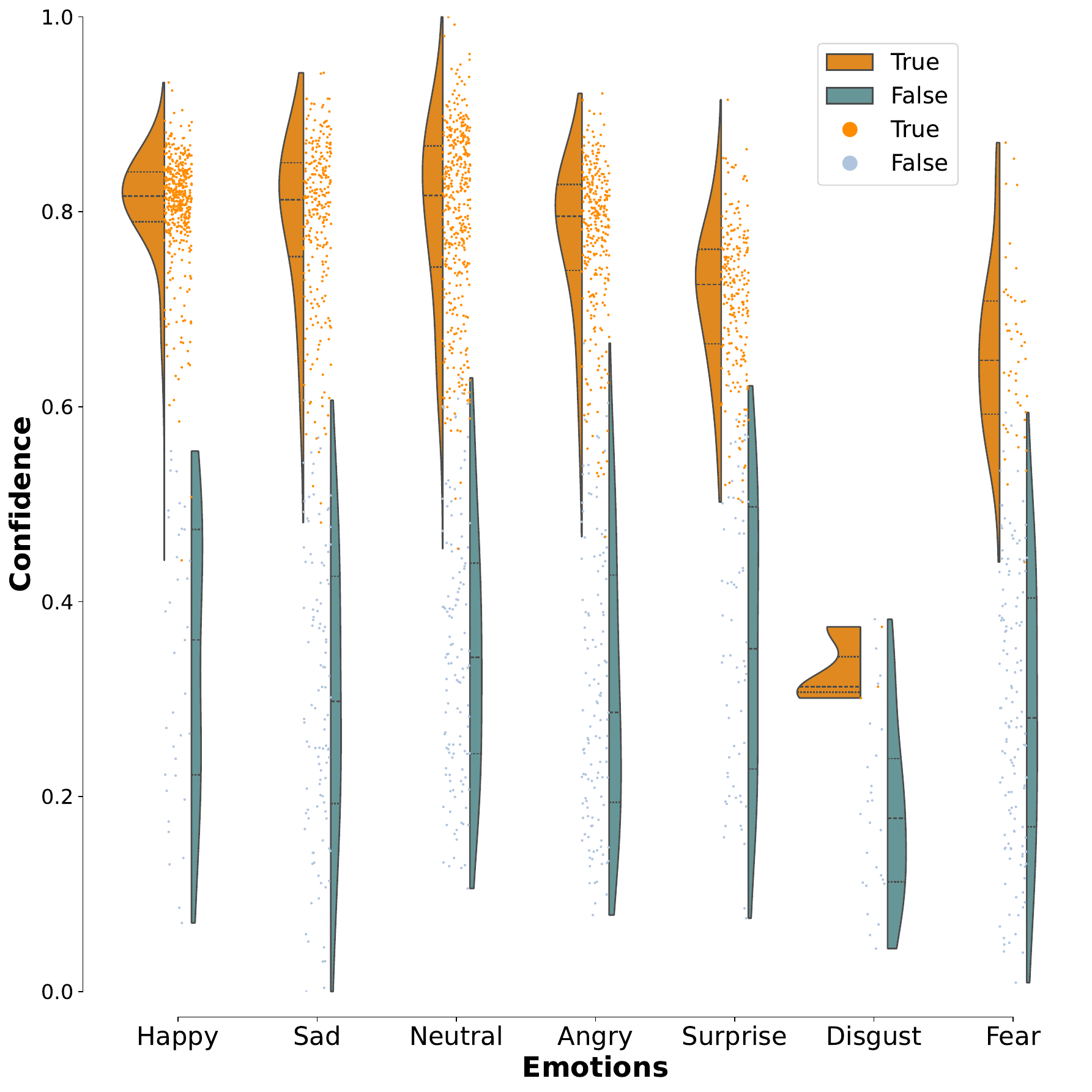}
    }
    \caption{Distribution of predicted per-class confidence on DFEW test set (1-th fold). Vanilla baselines and IFDD variants are involved. Sample points of different emotions and their kernel density are shown in different columns. Positive and negative samples are labeled by different colors, i.e., orange for positive ones and green for negative ones.\label{fig:CLSConf} }
\end{figure}

\begin{table*}[t]
    \centering    
    \begin{tabular}{ccccc}
            \toprule
            \multirow{2}{*}{Network} & \multirow{2}{*}{Backbone} & \multirow{2}{*}{Head} & \multicolumn{2}{c}{Efficiency} \\
            \cmidrule(l){4-5}
            & & & Params (M) & FLOPs (G) \\
            \midrule
            \multicolumn{1}{c}{2DCNN Baseline} & \multirow{2}{*}{MobileNetV2} & Vanilla (MLP) & 1.55 & 3.73 \\
            \cmidrule(l){1-1}\cmidrule(l){3-5}
            \multicolumn{1}{c}{IFDD-2DCNN} & & IFDD & 1.18 & 3.86 \\
            \midrule
            \multicolumn{1}{c}{3DViT Baseline} & \multirow{2}{*}{MViT} & Vanilla (MLP) & 23.41 & 32.83 \\
            \cmidrule(l){1-1}\cmidrule(l){3-5}
            \multicolumn{1}{c}{IFDD-3DViT} & & IFDD & 37.44 & 36.01 \\
            \bottomrule
    \end{tabular}
    \caption{Efficiency Comparison with Baselines.\label{Tab. Efficiency}}
\end{table*}

\begin{table*}[!ht]
    \centering
    \renewcommand{\arraystretch}{1.2}
    \resizebox{0.85\linewidth}{!}{
        \begin{tabular}{c|cccccccccc}
            \hline
            \multirow{3}{*}{IFDD Variants} & \multicolumn{10}{c}{DFEW Dataset} \\ \cline{2-11} 
             & \multicolumn{2}{c|}{fold 1} & \multicolumn{2}{c|}{fold 2} & \multicolumn{2}{c|}{fold 3} & \multicolumn{2}{c|}{fold 4} & \multicolumn{2}{c}{fold 5} \\ \cline{2-11} 
             & UAR & \multicolumn{1}{c|}{WAR} & UAR & \multicolumn{1}{c|}{WAR} & UAR & \multicolumn{1}{c|}{WAR} & UAR & \multicolumn{1}{c|}{WAR} & UAR & WAR \\ \hline
            IFDD-2DCNN & 56.02 & \multicolumn{1}{c|}{70.03} & 55.23 & \multicolumn{1}{c|}{67.39} & 56.74 & \multicolumn{1}{c|}{70.93} & 57.68 & \multicolumn{1}{c|}{70.27} & 57.96 & 71.49 \\ \hline
            IFDD-3DViT & \multicolumn{1}{l}{59.11} & \multicolumn{1}{c|}{73.43} & 59.50 & \multicolumn{1}{c|}{71.89} & 62.37 & \multicolumn{1}{c|}{74.62} & 63.38 & \multicolumn{1}{c|}{74.12} & 61.60 & 75.06 \\ \hline
            \multirow{2}{*}{IFDD Variants} & \multicolumn{10}{c}{MAFW Dataset} \\ \cline{2-11} 
             & \multicolumn{2}{c|}{fold 1} & \multicolumn{2}{c|}{fold 2} & \multicolumn{2}{c|}{fold 3} & \multicolumn{2}{c|}{fold 4} & \multicolumn{2}{c}{fold 5} \\ \hline
            Metrics & UAR & \multicolumn{1}{c|}{WAR} & UAR & \multicolumn{1}{c|}{WAR} & UAR & \multicolumn{1}{c|}{WAR} & UAR & \multicolumn{1}{c|}{WAR} & UAR & WAR \\ \hline
            IFDD-2DCNN & 29.09 & \multicolumn{1}{c|}{42.38} & 31.44 & \multicolumn{1}{c|}{46.60} & 34.90 & \multicolumn{1}{c|}{50.24} & 37.53 & \multicolumn{1}{c|}{54.20} & 35.45 & 51.30 \\ \hline
            IFDD-3DViT & 31.46 & \multicolumn{1}{c|}{47.09} & 37.83 & \multicolumn{1}{c|}{51.06} & 42.34 & \multicolumn{1}{c|}{57.39} & 43.98 & \multicolumn{1}{c|}{57.81} & 40.95 & 56.25 \\ \hline
            \end{tabular}
    }
    \caption{5-fold Results for DFEW and MAFW datasets.\label{Tab. 5-fold}}
\end{table*}

\begin{table*}[htb]
    \center
    \caption{Comparisons with self-supervised methods on DFEW dataset. - means not reported in the corresponding paper.\label{Tab. DFEW_selfsup}}
    \renewcommand{\arraystretch}{1.2}
    \resizebox{\linewidth}{!}{
    \begin{tabular}{ccccccccccccccc}
    \toprule
    \multirow{2}{*}{Methods} & \multirow{2}{*}{Network Type} & \multirow{2}{*}{External Sources} & \multicolumn{2}{c}{Efficiency} & \multicolumn{7}{c}{Accuracy of Each Emotion $(\%)$} & \multicolumn{2}{c}{Metrics $(\%)$}\\
    \cmidrule(l){4-5} \cmidrule(l){6-12} \cmidrule(l){13-14}
    & & & Params (M) & FLOPs (G)  & \emph{Happy} & \emph{Sad} & \emph{Neutral} & \emph{Angry} & \emph{Surprise} & \emph{Disgust} & \emph{Fear} & UAR & WAR 
    \\ 
    \midrule 
    \multicolumn{4}{l}{ \emph{Supervised models}} \\
    \hline
    IFDD-2DCNN (ours) & 2D CNN + IFDD & w/o  & 1.18  & 3.86 & 90.80 & 71.24 & 69.48 & 70.80 & 65.99 & 0.00 & 28.73 & 56.72 & 70.01 \\
    IFDD-3DViT (ours) & 3D ViT + IFDD & w/o & 37.44 & 36.01 & 94.07 & 75.99 & 74.16 & 74.02 & 69.39 & 10.34 & 30.39 & 61.19 & 73.82 \\
    \hline
    \multicolumn{4}{l}{\emph{Self-supervised models with large-scale external source}} \\
    \hline
    DFER-CLIP\cite{DFER-CLIP} & CLIP + Temporal Adaption & WIT400M for CLIP & 90  & - & - & - & - & - & - & - & - & 59.61 & 71.25 \\
    MAE-DFER\cite{MAEDFER} & 3D ViT + MLP for Token & VoxCeleb2 & 85  & 50 & 92.92 & 77.46 & 74.56 & 76.94 & 60.99 & 18.62 & 42.35 & 63.41 & 74.43 \\
    \bottomrule
    \end{tabular}}
\end{table*}

\subsection{Allowable Range}
Recalling that the range of learned offset scales $\textbf{A}_S$ and $\textbf{A}_D$ is $(-1,1)$, the actual offsets are obtained by multiplying $\textbf{A}_S$ and $\textbf{A}_D$ with a restraint ratio $L$.
Then the content-aware splitting indices $\textbf{I}_S$ and $\textbf{I}_D$ are generated as the element-wise summation of offsets and initial indices.
We evaluate the impacts of different restraints $L$ on the offset range and different initial indices.
Table. \ref{Tab. OffsetRange} presents two observations. 
Firstly, larger allowable range performs better than the smaller one with a $0.79\%$ improvement in WAR, which indicates that more flexibility for index sampling benefits the disentanglement process.
Secondly, on-the-fly generation for splitting indices starting from even-odd indices outperforms the one starting from the temporal midpoint.
That is, initial status from even-odd splitting indices provide larger temporal receptive field than the one from temporal midpoint.

\subsection{Aggregation or Disentanglement first?}
The proposed LADM first aggregates coarsely splitting features to obtain global context features by an updater.
Then LADM disentangles emotion-relevant dynamic features from global context by a predictor.
In this way, LADM places updater ahead of predictor, which is in contrast with the vanilla lifting scheme.
To explore the impact of the order, we evaluate different orders of aggregation and disentanglement for LADM.
As shown in Table \ref{Tab. UPOrder}, conduct aggregation operation ahead of disentanglement achieves higher accuracy with $1.29\%$ increment.
It indicates that aggregating first provide more fine-grained global context, which will benefit the subsequent disentanglement operation.

\section{D. Additional Information for Visualization}
Here we provide supplementary information for Figure 2 in the main paper, especially implementation details.

\subsection{Gradient Attention of $\{Y_D,Y_S\}$}
For Figure 2(a), gradients from target emotion and features are extracted after LayerNorm Layer in LADM module.
Average pooling operation for spatial dimensions is conducted on gradients to get channel-wise weighted scores for features.
In this way, the attention maps of emotion-related dynamic features $Y_D$ and global context features $Y_S$ are visualized along temporal dimension.
The temporal size of $Y_D$ and $Y_S$ is $\frac{T}{2} = 4$ in LADM, and the original clips with temporal size = 16 are downsampled in temporal dimension by stride = 4 for visualization.
Figure 2(a) has demonstrated that IFDD can learn to disentangle dynamic features by focusing on emotion-related regions, especially decoupling in an implicit way without the guidance of external sources such as facial region labels or landmarks.

\subsection{Distribution of Classification Features}
As for Figure 2(b), we extract 768-dimension feature vectors before the last linear layer, and visualize them in a 2-D space by t-SNE.
The perplexity of t-SNE is set to 50, while the maximum number of iterations is set to 1000.
Figure 2(b) has shown that the disentangled facial dynamic features by IFDD have stronger correlations with target emotion-related expression.
Notably, the stepwise participation of ISSM and LADM progressively widens the distribution gaps between high-level features of different emotions, which proves that ISSM and LADM both contribute to the discrimination capacity of the model.

\section{E. Detailed Comparison with Baselines}
To further demonstrate the effectiveness of IFDD, we conduct quantitative and qualitative analysis for comprehensive comparison with CNN and ViT baselines.
Here we briefly introduce the prediction heads of these two baselines.
The prediction head for 2D-CNN baseline (MobileNetV2) concatenates features across temporal dimension and then directly flatten features into 1-D vector, followed by an MLP layer for final classification. 
Similarly, the prediction head for 3D-ViT baseline (MViT) also uses an MLP layer to compress the class token of MViT for final classification.

\subsection{Per-emotion Accuracy Comparison}
Tab. \ref{Tab. OverviewAblation} and Figure \ref{fig:Radar} indicate a broad improvement by our methods across most emotions, especially $36.37\%$/$28.00\%$ by IFDD-2DCNN/3DViT for sad emotion which is easy to be confused with neural emotion.
However, the accuracy of \emph{Disgust} and \emph{Fear} emotion is low, since DFEW dataset has severe long-tailed distribution issue for different emotions.

\subsection{Extra Computational Cost}
As for efficiency comparison shown in Table \ref{Tab. Efficiency}, CNN and ViT baselines equipped with our proposed IFDD have incurred $3.49\%$ and $9.69\%$ increments in extra computational cost, which is quite tolerable.
Notably, since the 128-dimension features utilized by IFDD-2DCNN have lower embedding dimension than the 768-dimension ones by IFDD-3DViT, IFDD-2DCNN even has a lower parameter count compared to its vanilla baseline, demonstrating the efficiency of IFDD for low-dimensional data.

\subsection{Distribution of Confidence Scores}
To further discuss the superior of our methods to baselines, we visualize the confidence scores of predictions for IFDD.
The predicted confidence scores of specific emotion are placed along the y-axis in Figure \ref{fig:CLSConf}, with positive and negative samples distinguished by different colors.
When compared to baselines, we can observe that both IFDD-2DCNN and IFDD-3DViT considerably refine the discriminative property by widening gap between positive and negative samples.
Besides, IFDD-3DViT presents more distinctive confidence scores.

\section{F. 5-Fold Cross-Validation Results}
Recall that we report the average UAR and WAR values of 5-fold results for DFEW and MAFW datasets as the final results in the main paper, following the experimental settings of DFEW and MAFW.
As a supplement, here we provide the corresponding 5-fold cross-validation results, as shown in Table \ref{Tab. 5-fold}.
UAR and WAR metrics for DFEW and MAFW datasets are evaluated following 5-fold cross-validation settings provided by these two datasets.

\section{G. Additional Discussions}
\subsection{Compared with Self-supervised Methods}
As stated in the main manuscript, we do not compare with self-supervised methods for fair comparison.
Here we further discuss the performance difference between IFDD and state-of-the-art self-supervised DFER methods including MAE-DFER and DFER-CLIP.

As shown in Table \ref{Tab. DFEW_selfsup}, IFDD-3DViT outperforms DFER-CLIP, but is surpassed by MAE-DFER on DFEW dataset.
Despite the absence of large-scale pre-training and external data, IFDD achieves comparable accuracy to these methods.


\end{document}